\documentclass[10pt,twocolumn,letterpaper]{article}

\usepackage{iccv}
\usepackage{times}
\usepackage{epsfig}
\usepackage{subcaption}
\usepackage{multirow}
\usepackage{graphicx}
\usepackage{amsmath}
\usepackage{amssymb}
\usepackage{booktabs}
\usepackage{algorithm}
\usepackage{algorithmic}
\usepackage{pifont}

\usepackage[pagebackref=true,breaklinks=true,letterpaper=true,colorlinks,bookmarks=false]{hyperref}

\DeclareMathOperator{\argmin}{arg\,min}

\iccvfinalcopy % *** Uncomment this line for the final submission

\def\iccvPaperID{9823} % *** Enter the ICCV Paper ID here

\ificcvfinal\fi

\begin{document}

%%%%%%%%% TITLE
\title{Communication-efficient Federated Learning with Single-Step Synthetic Features Compressor for Faster Convergence}

\author{Yuhao Zhou$^1$, Mingjia Shi$^1$, Yuanxi Li$^2$, Qing Ye$^1$, Yanan Sun$^1$, Jiancheng Lv$^1$\\
$^1$~Sichuan University\\
$^2$~University of Illinois at Urbana-Champaign\\
% {\tt\small \{sooptq, 3101ihs\}@gmail.com, \{yeqing, ysun, lvjiancheng\}@scu.edu.cn}
% \and
{\tt\small \{sooptq, 3101ihs\}@gmail.com, yuanxi3@illinois.edu, \{yeqing, ysun, lvjiancheng\}@scu.edu.cn}
}

\maketitle
\ificcvfinal\thispagestyle{empty}\fi

%%%%%%%%% ABSTRACT
\begin{abstract}
   Reducing communication overhead in federated learning (FL) is challenging but crucial for large-scale distributed privacy-preserving machine learning. While methods utilizing sparsification or other techniques can largely reduce the communication overhead, the convergence rate is also greatly compromised. In this paper, we propose a novel method named Single-Step Synthetic Features Compressor (3SFC) to achieve communication-efficient FL by directly constructing a tiny synthetic dataset containing synthetic features based on raw gradients. Therefore, 3SFC can achieve an extremely low compression rate when the constructed synthetic dataset contains only one data sample. Additionally, the compressing phase of 3SFC utilizes a similarity-based objective function so that it can be optimized with just one step, considerably improving its performance and robustness. To minimize the compressing error, error feedback (EF) is also incorporated into 3SFC. Experiments on multiple datasets and models suggest that 3SFC has significantly better convergence rates compared to competing methods with lower compression rates (i.e., up to 0.02\%). Furthermore, ablation studies and visualizations show that 3SFC can carry more information than competing methods for every communication round, further validating its effectiveness.
\end{abstract}

%%%%%%%%% BODY TEXT
\section{Introduction}
Until now, federated learning~\cite{mcmahan2017communication} (FL) is deemed as one of the most promising distributed techniques~\cite{dean2012large,ben2019demystifying} to tackle the isolated data island problem with privacy guarantees. However, the training process of FL involves frequent exchanging of model parameters between central servers and participating clients, which is becoming increasingly expensive, especially considering the rapid growth of model size today~\cite{brown2020language,kaplan2020scaling,fedus2021switch}. Moreover, participating clients of FL typically operate at unreliable and limited network connection rates compared to data centers~\cite{kairouz2021advances}, further hindering the large-scale deployments of FL. Consequently, communication is becoming the primary bottleneck for flexible FL at the scale~\cite{bonawitz2019towards}.

To explore possible approaches for reducing communication overhead in FL, various methods have been proposed targeting different objectives. The work in \cite{strom2015scalable,lin2017deep} applied top-$k$ sparsification to the gradients so that only the most important information is transmitted at each epoch. Moreover, Wangni~\cite{wangni2018gradient} reported that using top-$k$ sparsification with error feedback (EF), the communication overhead of ResNet-50~\cite{he2016deep} trained on ImageNet~\cite{russakovsky2015imagenet} could be reduced by 99.6\% while maintaining nearly the same model accuracy. On the other hand, The work in \cite{alistarh2017qsgd,bernstein2018signsgd} employed quantification to represent gradients by a lower precision data type with a considerably smaller size. Later Karimireddy~\cite{karimireddy2019error} introduced error feedback to quantification as well, substantially improving the rate of convergence. In \cite{goetz2020federated}, instead of gradients, several data samples distilled from the full training dataset were transmitted as they are much smaller than the gradients, and they can produce similar gradients through back-propagation. More recently, Li~\cite{li2021communication} and Wu~\cite{wu2022communication} proposed compressing and decompressing communication data using compressed sensing and knowledge distillation, respectively.
\begin{figure}[tb]
    \centering
    \includegraphics[width=0.35\textwidth]{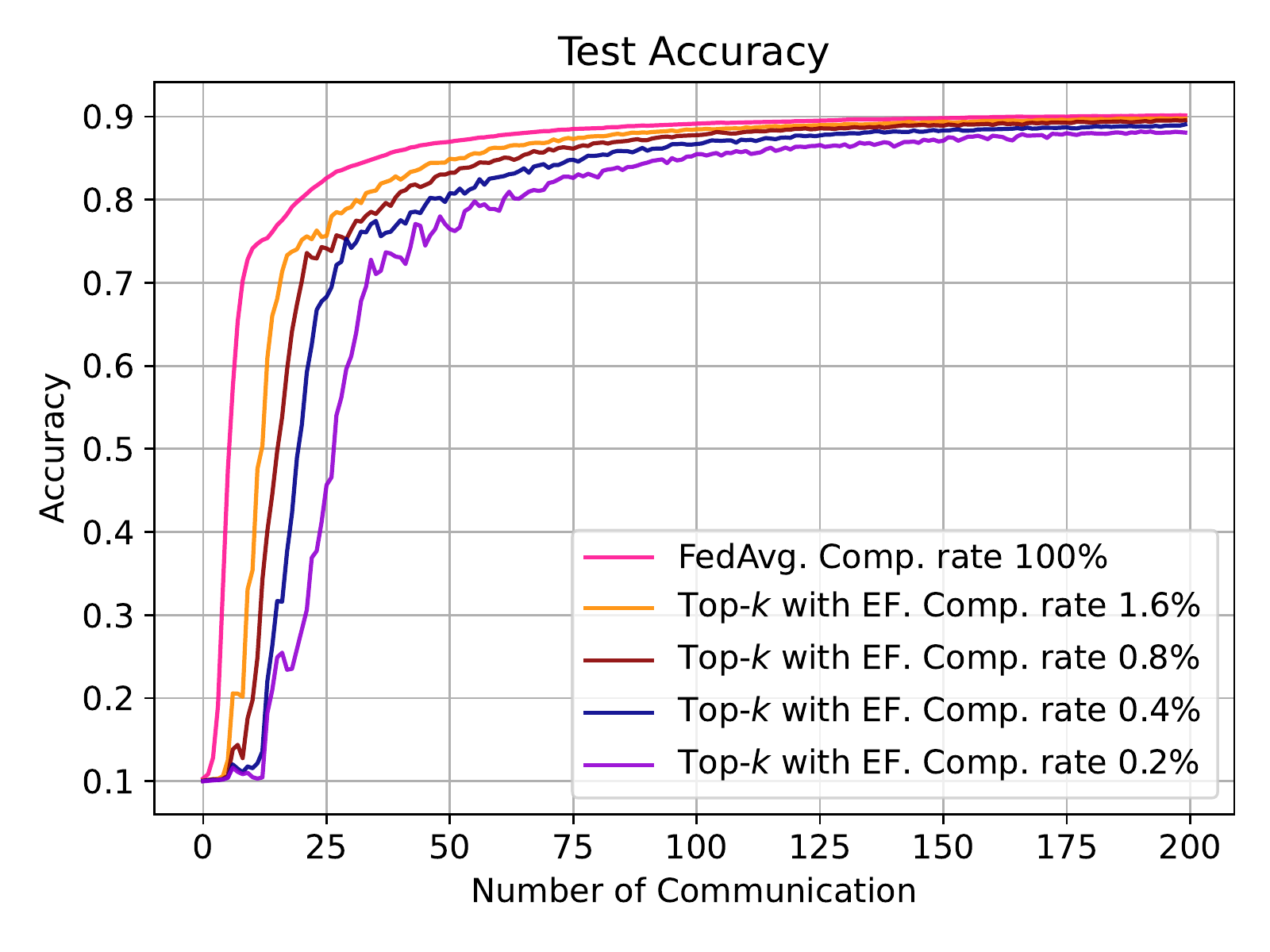}
    \caption{Test accuracy of MLP (199,210 parameters) trained on non-i.i.d. MNIST dataset with 20 clients. The rate of convergence reduces as the compression rate decreases.}
    \label{fig:pre-degraded-convergence-rate}
\end{figure}

While the methods mentioned above have been proven useful in reducing communication overhead reduction, empirical evidence indicates that they both suffer from degraded model convergence rates. This means that the model is expected to converge much slower with a smaller compression rate, as demonstrated in Figure ~\ref{fig:pre-degraded-convergence-rate}. Here, the compression rate is defined as Equation~\ref{eq:define-compression-rate}. In this paper, we propose a single-step synthetic features compressor (3SFC), to boost the convergence rate during training and carry more information under a limited communication budget. Instead of transmitting raw gradients directly, 3SFC first constructs a tiny synthetic dataset for the FL model. Then a scaling coefficient is calculated to minimize the compression error. Finally, the constructed synthetic dataset and the scale coefficient are transmitted to the server. In addition, the error feedback~\cite{stich2018sparsified} is also incorporated into 3SFC to further minimize the overall compression error.
\begin{equation}
    \text{Comp. Rate} = \frac{\text{Comp. Size}}{\text{Uncomp. Size}} = \frac{1}{\text{Comp. Ratio}}.
    \label{eq:define-compression-rate}
\end{equation}

Our contributions can be summarized as following:

\begin{enumerate}
    \item Instead of transmitting gradients employed by most existing compression methods, 3SFC only transmits a tiny set of model inputs and labels, which is independent of the model architecture. Consequently, 3SFC can achieve an extremely low compression rate.
    \item A similarity-based objective function is employed to construct synthetic inputs and labels, which drastically lowers the time and space complexity and improves the performance and robustness of 3SFC. Moreover, the error feedback is incorporated into 3SFC to minimize the overall compressing error and therefore boost the convergence rate. These design choices make 3SFC an effective solution for achieveing communication-efficient FL while maintaining model accuracy.
    \item 3SFC can achieve a significantly better convergence rate compared to competing methods under the same and even lower communication budget (\textit{i.e.}, up to a compression ratio of 3600$\times$). Ablation study and other visualizations further validate the efficiency of 3SFC against other state-of-the-art works. The code is open-sourced for reproduction~\footnote{The source code is uploaded as the Supplementary Material, and will be open-sourced after publication.}.
\end{enumerate}
%------------------------------------------------------------------------
\section{Related Work}
\label{sec:related-work}
\textbf{Sparsification}: Methods in \cite{aji2017sparse,lin2017deep,wangni2018gradient} utilized sparsifiers to filter and send partial gradients to greatly reduce the communication overhead. Typical sparsifiers include random-$k$, top-$k$, etc. DGC~\cite{lin2017deep} and STC~\cite{sattler2019robust} are considered as current state-of-the-arts. Recently, Sahu~\cite{sahu2021rethinking} formally showed that top-$k$ is the communication-optimal sparsifier under a limited communication budget.

\textbf{Quantification}: Methods in \cite{bernstein2018signsgd,alistarh2017qsgd,seide20141} replaced the default 32-bit data type in Machine Learning (ML) training with the 8-bit or even 1-bit data type before communicating, and thereby reducing communication overhead. Compared to sparsification methods that can achieve a compression rate of 1/100 or even lower, quantification can at most achieve a compression rate of 1/32.
\begin{figure}[t]
    \centering
    \includegraphics[width=0.35\textwidth]{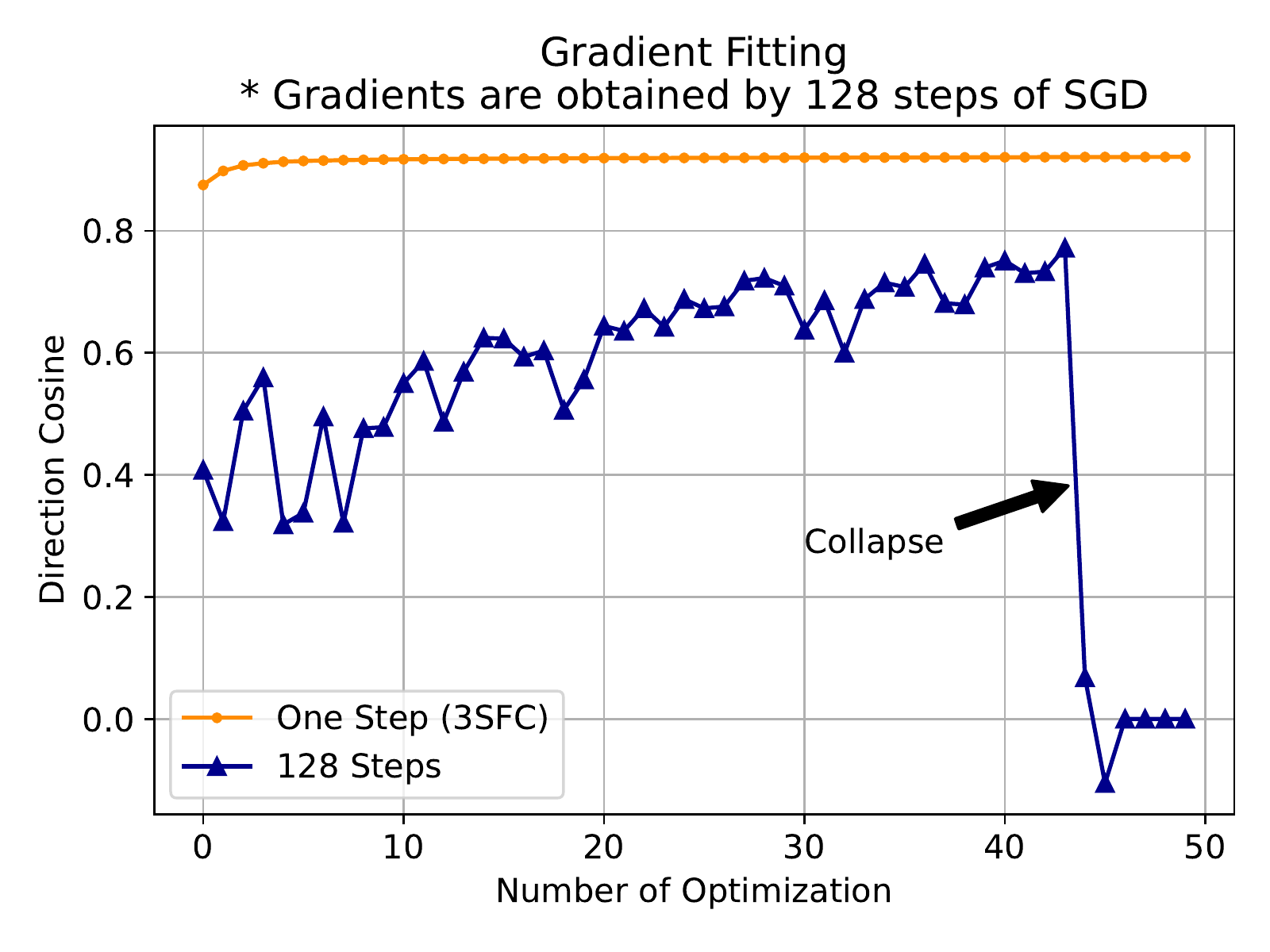}
    \caption{When trying to fit gradients obtained by 128 steps of SGD for 128 steps of simulation using the method in \protect\cite{goetz2020federated}, it should be perfectly fitted instead of collapsed. On the other hand, using 1 step of simulation (3SFC) requires less computation and storage but achieves significantly better fitting results.}
    \label{fig:pre-exp-cos}
\end{figure}
\begin{figure}[t]
    \centering
    \includegraphics[width=0.35\textwidth]{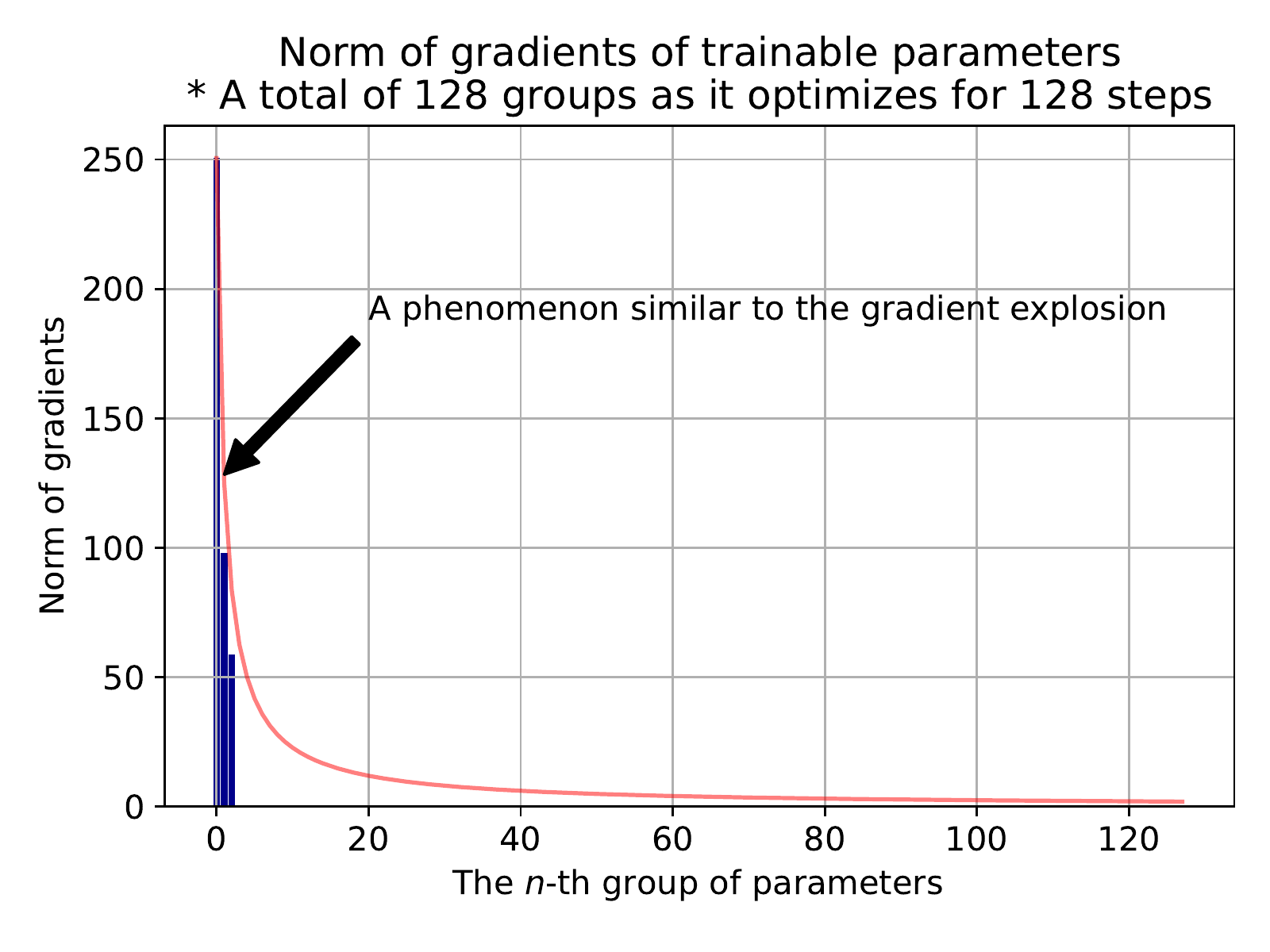}
    \caption{Before the collapse of the method in \protect\cite{goetz2020federated} in Figure~\ref{fig:pre-exp-cos}, the gradients of its trainable parameters exhibit a phenomenon similar to the gradient explosion, where the magnitude of gradients increases as they backpropagate from the 128-th to the first group of parameters. This could be a possible reason for the collapse.}
    \label{fig:pre-exp-grad}
\end{figure}

\textbf{Data distillation (Synthetic dataset) for FL}: Recent work \cite{goetz2020federated,hu2022fedsynth} have proposed using several synthetic data samples to represent gradients in FL. Since local models in FL are optimized for multiple steps locally, The synthesis process will first generate a synthetic dataset, then simulate optimizing its local model for multiple steps using the synthetic dataset, and finally, utilize the minimization of the $\ell_2$ distance between simulated and real model weights as its objective function to optimize the synthetic dataset. While the multiple steps of simulation is intuitive, empirical results suggest that it leads to not only a high level of time and space complexity for synthesizing (\textit{i.e.}, calculate gradients and store intermediate model weights multiple times), but also great instability and possible collapse, especially for relatively large models and datasets. Figure~\ref{fig:pre-exp-cos} demonstrates such a collapse. Moreover, gradients of trainable parameters before the collapse are visualized in Figure~\ref{fig:pre-exp-grad}, indicating that a phenomenon similar to the gradient explosion had occurred due to the multiple steps of simulation. Consequently, since previous work hardly converges under our experimental settings involving extremely low compression rate and relatively large models and datasets as Section~\ref{sec:experiments} demonstrated, they will not be compared in our formal experiments.

To alleviate the above-mentioned problems for data distillation to achieve both computation and communication efficient gradient compression, 3SFC employs a similarity-based objective function, instead of $\ell_2$-based objective function, to optimize the synthetic dataset. Empowered by the new objective function, 3SFC optimizes the synthetic dataset only once compared to dozens and hundreds by previous work, substantially enhancing the performance and stability under low compression rates and large models and datasets. Moreover, error feedback is introduced to the data distillation realm for the first time, further helping the optimization converge.
\begin{figure*}[!h]
    \centering
    \includegraphics[width=0.88\textwidth]{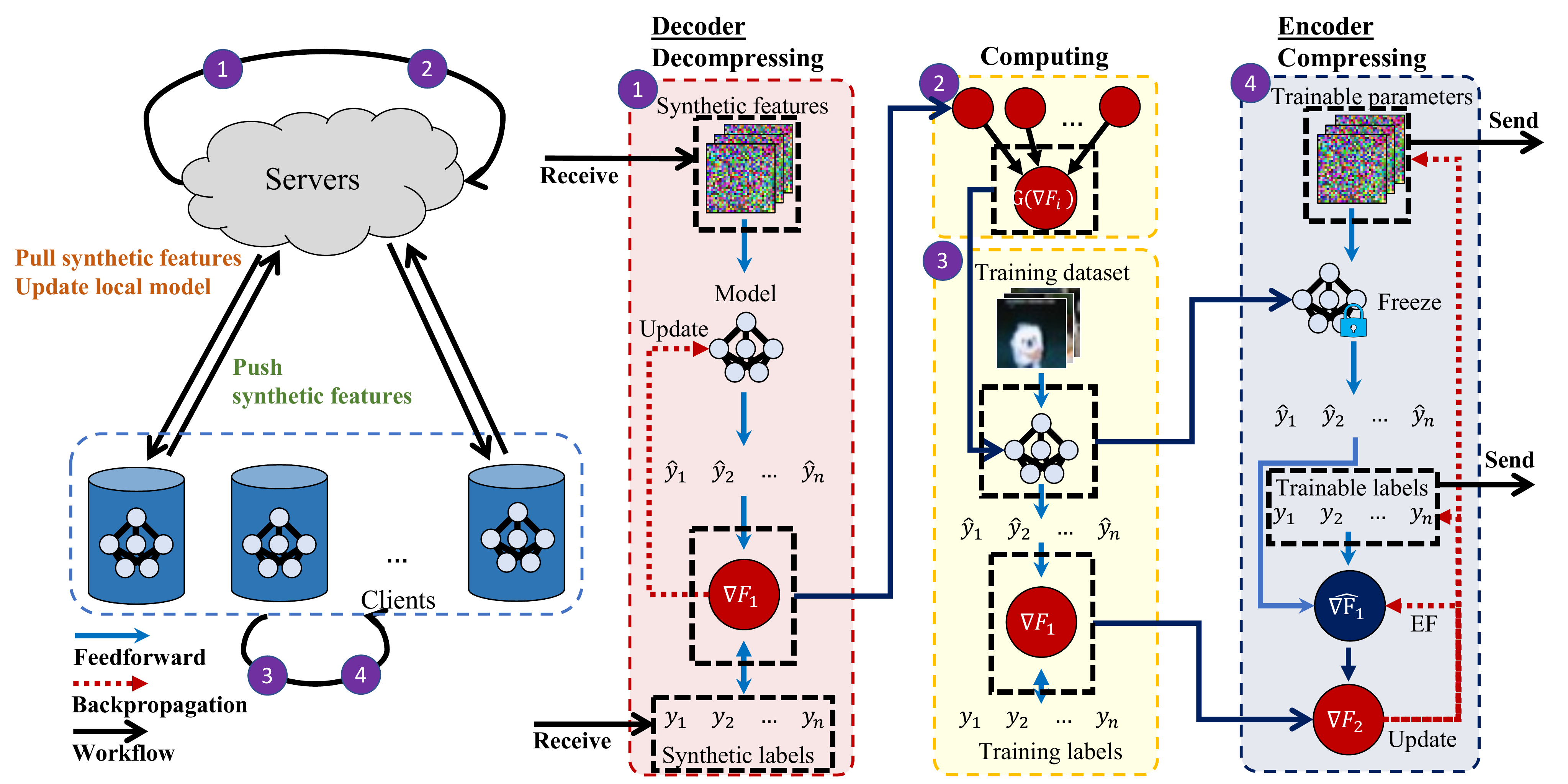}
    \caption{The general architecture of 3SFC. When compressing in \ding{185}, a set of trainable parameters and labels will first be fed into the frozen local model to calculate model gradients. Then, calculated model gradients will be compared with real model gradients to optimize the trainable parameters and labels (i.e., synthetic features). When decompressing in \ding{182}, simply feed the local model with the received synthetic inputs and labels and use the generated gradients to update the global model.}
    \label{fig:3sfc-arch}
\end{figure*}

\textbf{Others}: Li~\cite{li2021communication} proposed utilizing compressed sensing for communication data compressing and decompressing. Wu~\cite{wu2022communication} used knowledge distillation to distill the learned knowledge from the local model to a smaller model before communication.
%------------------------------------------------------------------------
\section{Problem Formulation}
Assume there are $N$ clients participating in a FL training process, where the $i$-th client has a local dataset $D_i$ that obeys distribution $P_i$, and a loss function $F_i(D_i, w_i)$ where $w_i$ is the weight of its model $M_i$. Note that in vanilla FL, all clients and servers share the same model architecture, \textit{i.e.,} $M_1 = M_2 = ... = M_N = M$. The objective of FL is to solve Equation~\ref{eq:fl-objective}:
\begin{equation}
    \min_{w \in \mathbb{R}^d} G(F_1(D_1, w), F_2(D_2, w), ..., F_N(D_N, w)),
    \label{eq:fl-objective}
\end{equation}
where $G(\cdot)$ is the linear aggregation function satisfying the sum of aggregation weights equals 1. Typical aggregation functions include arithmetic mean and weighted average based on $|D_i|$~\cite{mcmahan2017communication} where $|\cdot|$ is the size of the $\cdot$. The global model $w$ at the $t$-th communication round is updated by Equation~\ref{eq:eq-3}:
\begin{align}
\begin{split}
    w^{t+1} &= G(w_1^{t}, w_2^{t}, ..., w_N^{t})\\ 
    &= w^{t} - G(\boldsymbol{g}_1^{t}, \boldsymbol{g}_2^{t}, ..., \boldsymbol{g}_N^{t})
\end{split},
\label{eq:eq-3}
\end{align}
where $\boldsymbol{g}_i^t = w^t - w_i^t$ denotes the model weight differences after locally training for $K$ rounds, and can be seen as accumulated gradients. Generally, to reduce communication overhead, a compressor $\mathcal{C}$ is applied to each client's $\boldsymbol{g}_i^t$, so that the global model $w$ can be updated by Equation~\ref{eq:eq-4}:
\begin{equation}
    w^{t+1} = w^{t} - G(\mathcal{C}(\boldsymbol{g}_1^t), \mathcal{C}(\boldsymbol{g}_2^t), ..., \mathcal{C}(\boldsymbol{g}_N^t)).
\label{eq:eq-4}
\end{equation}

As a result, the objective of communication compressing can be modeled as Equation~\ref{eq:eq-5}:
\begin{equation}
    \mathcal{C}^* = \argmin || \mathcal{C}(\boldsymbol{g_i^t}) -\boldsymbol{g_i^t} ||^2 \text{ s.t. } ||\mathcal{C}(\boldsymbol{g_i^t})||_0 \leq B,
\label{eq:eq-5}
\end{equation}
where $B$ is the communication budget, constraining the maximum size of communication data at each communication round, and $||\cdot||$ measures the distances of $\cdot$. Moreover, letting $\boldsymbol{\epsilon}_i^t = ||\mathcal{C}(\boldsymbol{g_i^t}) - \boldsymbol{g_i^t}||$ denotes the compression error at time $t$, then the error feedback can be utilized to optimize this error term by adding it to the $\boldsymbol{g_i^{t+1}}$. Thus, with error feedback, the global model can be updated as Equation~\ref{eq:general-problem-optim}:
\begin{equation}
    \left\{
        \begin{split}
            w^{t+1} &= w^{t} - G(\mathcal{C}(\boldsymbol{g}_1^t + \boldsymbol{\epsilon}_1^{t}), \mathcal{C}(\boldsymbol{g}_2^t + \boldsymbol{\epsilon}_2^{t}), ..., \mathcal{C}(\boldsymbol{g}_N^t + \boldsymbol{\epsilon}_N^{t})), \\
            \boldsymbol{\epsilon}_i^{t+1} &= \boldsymbol{g}_i^t + \boldsymbol{\epsilon}_i^{t} - \mathcal{C}(\boldsymbol{g}_1^t + \boldsymbol{\epsilon}_1^{t}).
        \end{split}
    \right.
    \label{eq:general-problem-optim}
\end{equation}
%------------------------------------------------------------------------
\section{Our Approach}
The general architecture of 3SFC is illustrated in Figure~\ref{fig:3sfc-arch}. At each epoch, the $i$-th client first trains its local model using its local dataset. After training, accumulated gradients can be obtained by subtracting the global model weights from the latest local model weights. Then, the $i$-th client will utilize the encoder to compress the averaged gradients into a synthetic dataset $D_{syn,i}^t$ that fits the communication budget. When the compressed data is received by the server, the server will first decode the compressed data into accumulated gradients, and then it will aggregate the gradients and update the global model. As seen from Figure~\ref{fig:3sfc-arch}, in 3SFC, the compressor $\mathcal{C}$ consists of an encoder and a decoder, where the encoder is located on the clients and the decoder is placed on the servers.

\subsection{Encoder with error feedback}
The encoder in 3SFC is responsible for compressing $\boldsymbol{g}_i^t$ into a synthetic dataset $D_{syn,i}^t$ and a scaling coefficient $s_i^t$. The objective of the encoder can be described by Equation~\ref{eq:encoder-objective-1}:
\begin{equation}
    \left\{
    \begin{aligned}
    \begin{split}
    \min_{D_{syn,i}^t, s_i^t} ||s_i^t \nabla_{w^t} F_i(D_{syn,i}^t, w^t)  - \boldsymbol{g_i^t} - \boldsymbol{\epsilon}_i^t||^2 + \lambda {D_{syn,i}^t}^2 \\ \text{ s.t. } ||D_{syn,i}^t||_0 + 1 \leq B,
    \end{split}
    \end{aligned}
    \right.
\label{eq:encoder-objective-1}
\end{equation}
where $\boldsymbol{g_i^t}$ denotes the differences between the global model $w^t$ and its latest local model, \textit{i.e.,} $\boldsymbol{g_i^t} = w^t - w_i^{t}$ for clients. $\lambda {D_{syn,i}^t}^2$ is an $\ell_2$ regularization term to constrain the solution of $D_{syn,i}^t$ for better stability. Note that here the global model $w^t$ is passed into $F_i(\cdot)$ instead of $w_i^t$, because $w^t$ is the initial weight of every client's local optimization process at each epoch. Since $g_i^t + \epsilon_i^t$ is fixed, $s_i^t$ can be derived from $\nabla_{w^t} F_i(D_{syn,i}^t, w^t)$ as shown in Equation~\ref{eq:s-compute}:
\begin{align}
\begin{split}
    s_i^t &= \frac{||\boldsymbol{g_i^t} + \boldsymbol{\epsilon}_i^t||}{||\nabla_{w^t} F_i(D_{syn,i}^t, w^t)||} cos(\theta) \\ &= \frac{(\boldsymbol{g_i^t} + \boldsymbol{\epsilon}_i^t) \cdot \nabla_{w^t} F_i(D_{syn,i}^t, w^t)}{||\nabla_{w^t} F_i(D_{syn,i}^t, w^t)||^2},
\label{eq:s-compute}
\end{split}
\end{align}
where $\theta$ is the angle between two $\boldsymbol{g_i^t} + \boldsymbol{\epsilon}_i^t$ and $\nabla_{w^t} F_i(D_{syn,i}^t, w^t)$. Consequently, the objective described in Equation~\ref{eq:encoder-objective-1} is equivalent to the following optimization problem:
\begin{equation}
    \left\{
    \begin{aligned}
    \begin{split}
    \min_{D_{syn,i}^t} 1 - |\frac{\nabla_{w^t} F_i(D_{syn,i}^t, w^t) \cdot (\boldsymbol{g_i^t} + \boldsymbol{\epsilon}_i^t)}{||\nabla_{w^t} F_i(D_{syn,i}^t, w^t)|| ||\boldsymbol{g_i^t} + \boldsymbol{\epsilon}_i^t||}| + \lambda {D_{syn,i}^t}^2 \\\text{ s.t. } ||D_{syn,i}^t||_0 + 1 \leq B.
    \end{split}
    \end{aligned}
    \right.
\label{eq:dsyn-optim}
\end{equation}

Namely, The objective is to find a synthetic dataset $D_{syn}$ that produces gradients that are most similar to $\boldsymbol{g_i^t}$ in terms of the direction. After solving Equation~\ref{eq:dsyn-optim}, $s_i^t$ can be thus calculated by Equation~\ref{eq:s-compute} and the compression error $\boldsymbol{\epsilon}_i^t$ can be updated by Equation~\ref{eq:general-problem-optim}. Finally, $D_{syn,i}^t$ and $s_i^t$ will be uploaded to others to represent the local gradients of client $i$.
\begin{algorithm}[tb]
    \caption{3SFC}
    \label{alg:3sfc-algorithm}
    \textbf{Input}: global model $w^t$, local dataset $D_i$, learning rate $\eta_i$, accumulated gradient $\boldsymbol{\epsilon}_i^t$, regularization parameter $\lambda$\\
    \textbf{Parameter}: communication budget $B$, number of global epoch $E$, number of local iteration $K$, number of 3SFC iteration $S$, number of clients $N$, aggregation function $G$\\
    \textbf{Output}: global model $w^{t+1}$\\
    \textbf{Clients:}
    \begin{algorithmic}[1] %[1] enables line numbers
        \FOR{each client $i$ from $1$ to $N$ \textbf{in parallel}}
            \STATE initialize $D_{syn,i}^t$ where $||D_{syn,i}^t||_0 + 1 \leq B$
            \FOR{each local iteration $e$ from $1$ to $K$}
                \STATE $w_i^t = w_i^t - \eta_i \nabla_{w_i^t} F_i(D_i, w_i^t)$ \label{line:erm-optim}
            \ENDFOR
            \STATE $\boldsymbol{g}_i^t = w_i^t - w_i$
            \FOR{each $s$ from $1$ to $S$}
                \STATE $D_{syn,i}^t = D_{syn,i}^t - \eta_i \nabla_{D_{syn,i}^t} (1 - |\frac{\nabla_{w^t} F_i(D_{syn,i}^t, w^t) \cdot (\boldsymbol{g_i^t} + \boldsymbol{\epsilon}_i^t)}{||\nabla_{w^t} F_i(D_{syn,i}^t, w^t)|| ||\boldsymbol{g_i^t} + \boldsymbol{\epsilon}_i^t||}| + \lambda {D_{syn,i}^t}^2)$ \label{line:compression-optim}
            \ENDFOR
            \STATE $s_i^t = \frac{(\boldsymbol{g_i^t} + \boldsymbol{\epsilon}_i^t) \cdot \nabla_{w^t} F_i(D_{syn,i}^t, w^t)}{||\nabla_{w^t} F_i(D_{syn,i}^t, w^t)||^2}$
            \STATE $\boldsymbol{\epsilon}_i^{t+1} = \boldsymbol{\epsilon}_i^t + \boldsymbol{g}_i^t - \nabla_{w^t} F_i(D_{syn,i}^t, w^t)$
            \RETURN $D_{syn,i}^t$, $s_i^t$, $\boldsymbol{\epsilon}_i^{t+1}$
        \ENDFOR
    \end{algorithmic}
    \textbf{Servers:}
    \begin{algorithmic}[1] %[1] enables line numbers
        \FOR{each client $i$ from $1$ to $N$}
            \STATE receive $D_{syn,i}^t$, $s_i^t$
            \STATE $\boldsymbol{g}_i^t + \boldsymbol{\epsilon}_i^t = s_i^t \nabla_{w^t} F_i(D_{syn,i}^t, w^t)$
        \ENDFOR
        \STATE $w^{t+1} = w^{t} - G(\boldsymbol{g}_1^t + \boldsymbol{\epsilon}_1^t, \boldsymbol{g}_2^t + \boldsymbol{\epsilon}_2^t, ..., \boldsymbol{g}_N^t + \boldsymbol{\epsilon}_N^t)$
        \RETURN $w^{t+1}$
    \end{algorithmic}
\end{algorithm}

\subsection{Decoder}
After receiving $D_{syn,i}^t$ and $s_i^t$ from others, the decoder at server $j$ will attempt to reconstruct the gradients for local model updating by the following equation:
\begin{equation}
    \boldsymbol{g_i^t} + \boldsymbol{\epsilon}_i^t = s_i^t \nabla_{w^t} F_j(D_{syn,i}^t, w^t).
\end{equation}

Note that the success of the reconstruction depends on the assumption that the server $j$ has access to the global model $w^t$ and $F_i(\cdot) = F_j(\cdot)$, which can be easily satisfied. Finally, for servers, following Equation~\ref{eq:general-problem-optim}, the global model of server $j$ can be updated accordingly.

\subsection{Algorithm and complexity analysis}
The pseudocode of 3SFC is presented in Algorithm~\ref{alg:3sfc-algorithm}. In 3SFC, during the training process, clients will solve two optimization problems instead of one compared to the vanilla FL method FedAvg~\cite{mcmahan2017communication}: the empirical risk minimization problem on the local dataset (Line~\ref{line:erm-optim}) and Equation~\ref{eq:dsyn-optim} for compression (Line~\ref{line:compression-optim}). The solvers to these two problems are not nested, meaning the time complexity of 3SFC equals $\mathcal{O}(NE(K + S))$. In terms of the space complexity, 3SFC additionally stores the $w^t$, $D_{syn,i}^t$, $s_i^t$ and $\boldsymbol{\epsilon}_i^t$, which are all fixed size parameters. Hence, 3SFC shares the same space complexity, $\mathcal{O}(N)$, with FedAvg as well.
\begin{figure}
     \centering
     \begin{subfigure}[tb]{0.23\textwidth}
         \centering
         \includegraphics[width=\textwidth]{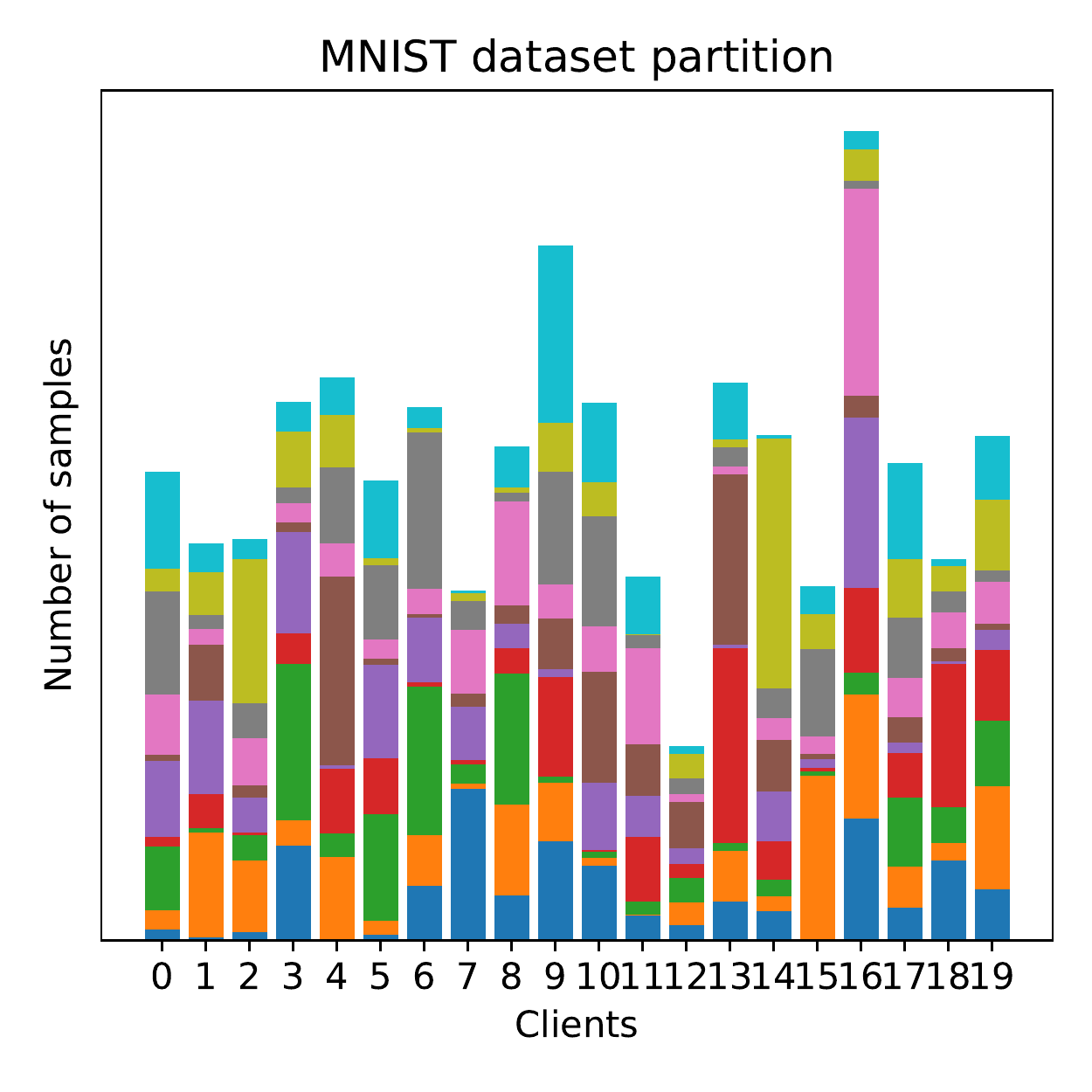}
     \end{subfigure}
     \hfill
     \begin{subfigure}[tb]{0.23\textwidth}
         \centering
         \includegraphics[width=\textwidth]{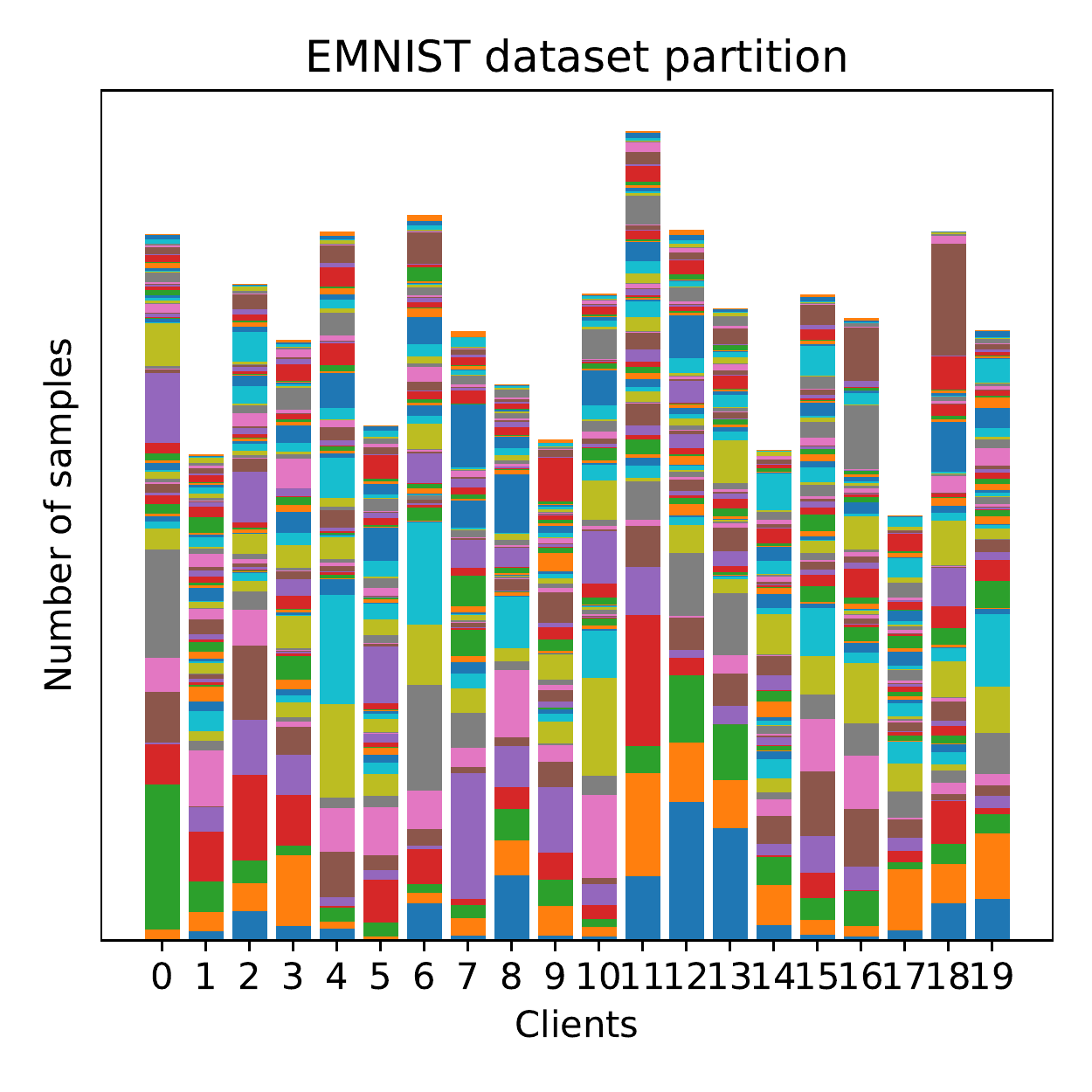}
     \end{subfigure}
     \\
     \begin{subfigure}[tb]{0.23\textwidth}
         \centering
         \includegraphics[width=\textwidth]{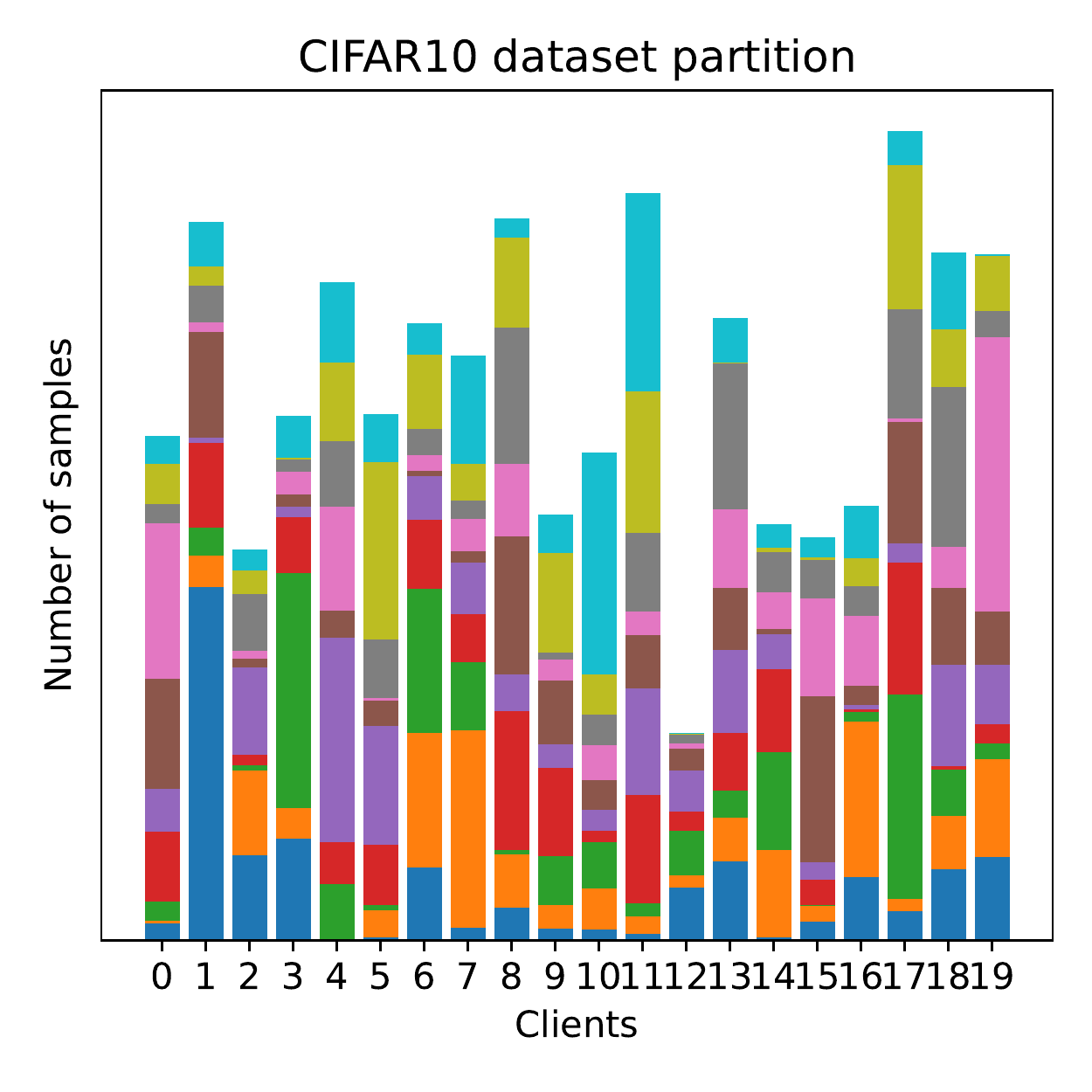}
     \end{subfigure}
     \hfill
     \begin{subfigure}[tb]{0.23\textwidth}
         \centering
         \includegraphics[width=\textwidth]{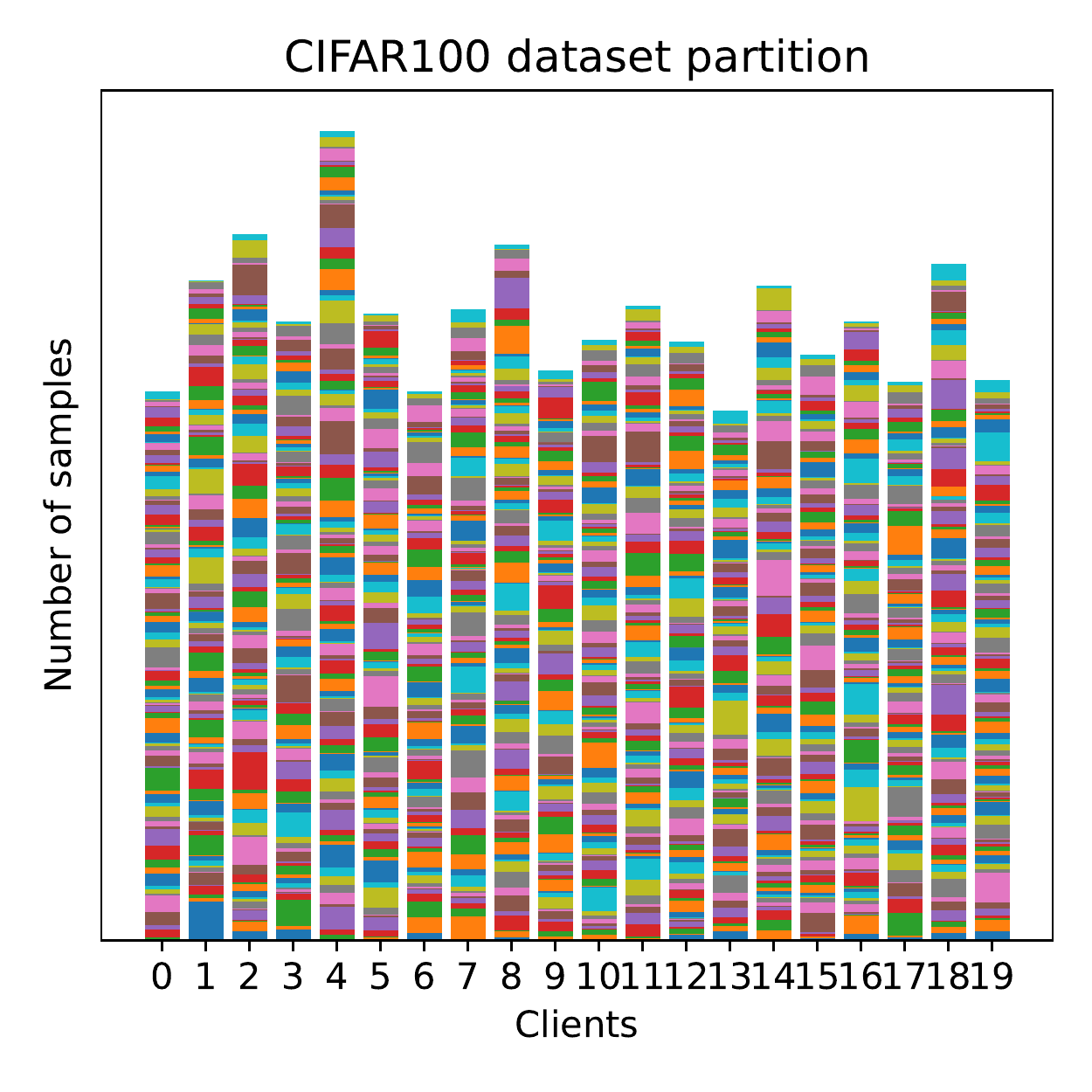}
     \end{subfigure}
        \caption{Illustration of our manual dataset partitions for 20 clients based on the Dirichlet distribution. Each bar represent a client, and different segments with different colors of a bar represents different labels. As can be seen, different clients have different dataset sizes and dataset distributions, and some clients only have some of the labels.}
        \label{fig:partition-illustration}
\end{figure}
% \begin{table}[t]
%   \centering
%   \resizebox{\linewidth}{!}{%
%     \begin{tabular}{lrrrrr}
%     \toprule
%     Dataset+Model & \multicolumn{1}{l}{FedAvg} & \multicolumn{1}{l}{DGC} & \multicolumn{1}{l}{signSGD} & \multicolumn{1}{l}{STC} & \multicolumn{1}{l}{3SFC} \\
%     \midrule
%     MNIST+MLP & 1.0$\times$     & 250.0$\times$   & 32.0$\times$    & 32.0$\times$    & 250.0$\times$ \\
%     EMNIST+MLP & 1.0$\times$     & 250.0$\times$   & 32.0$\times$    & 32.0$\times$    & 250.0$\times$ \\
%     FMNIST+MLP & 1.0$\times$     & 250.0$\times$   & 32.0$\times$    & 32.0$\times$    & 250.0$\times$ \\
%     FMNIST+Mnistnet & 1.0$\times$     & 1333.3$\times$   & 32.0$\times$    & 32.0$\times$    & 1333.3$\times$ \\
%     Cifar10+Convnet & 1.0$\times$     & 10.4$\times$ & 32.0$\times$    & 32.0$\times$    & 10.4$\times$ \\
%     Cifar10+Resnet & 1.0$\times$     & 3571.4$\times$ & 32.0$\times$    & 32.0$\times$    & 3571.4$\times$ \\
%     Cifar10+Regnet & 1.0$\times$     & 757.6$\times$ & 32.0$\times$    & 32.0$\times$    & 757.6$\times$ \\
%     Cifar100+ResNet & 1.0$\times$     & 3571.4$\times$ & 32.0$\times$    & 32.0$\times$    & 3571.4$\times$ \\
%     Cifar100+RegNet & 1.0$\times$     & 769.2$\times$ & 32.0$\times$    & 32.0$\times$    & 769.2$\times$ \\
%     \bottomrule
%     \end{tabular}%
%     }
%   \caption{Comparison of compression ratio. DGC is set to be the same as 3SFC, while signSGD and STC are set to $32$ due to data type limitation.}
%   \label{tab:comp-ratio}%
% \end{table}%

%------------------------------------------------------------------------
\section{Experiments}
\label{sec:experiments}
\textbf{Datasets:} Following the conventions of the community~\cite{sattler2019robust,zhou2021communication,bernstein2018signsgd}, five datasets including MNIST~\cite{deng2012mnist}, FMNIST~\cite{xiao2017fashion}, EMNIST~\cite{cohen2017emnist}, Cifar10 and Cifar100~\cite{krizhevsky2009learning} are used in the experiments. To simulate the Non-i.i.d. characteristic, all datasets are manually partitioned into multiple subsets based on the Dirichlet distribution, which is commonly used in the FL setting~\cite{wang2020tackling,li2022federated}. Figure~\ref{fig:partition-illustration} illustrates our partitions. As can be seen, different clients own different datasets in terms of both quantity and category.
\begin{table}[tb]
  \centering
  \resizebox{0.7\linewidth}{!}{%
    \begin{tabular}{cccc}
    \toprule
    \multirow{2}[2]{*}{Dataset+Model} & \multicolumn{1}{c}{\multirow{2}[2]{*}{FedAvg (1$\times$)}} & \multicolumn{2}{c}{FedSynth} \\
      &       & \multicolumn{1}{c}{1$\times$} & \multicolumn{1}{c}{250$\times$} \\
    \midrule
    MNIST+MLP & 0.9017 & 0.9017   & 0.1359 \\
    EMNIST+MLP & 0.6108 & 0.6108   & 0.0192 \\
    FMNIST+MLP & 0.8183 & 0.8183   & 0.1216 \\
    FMNIST+Mnistnet & 0.8573 & 0.8573   & 0.1318 \\
    \bottomrule
    \end{tabular}%
    }
  \caption{Test accuracies of FedSynth in our preliminary experiments with 10 clients after 200 epochs of training. As can be seen, the model is barely optimized (as discussed in Section~\ref{sec:related-work}) with FedSynth with an extremely high compression ratio, while other methods like 3SFC and DGC achieve much higher performances as Table~\ref{tab:accuracy-compare} illustrated. Consequently, FedSynth is not compared with 3SFC in the latter experiments. These results validate our observations described in Section~\ref{sec:related-work}.}
  \label{tab:fedsynth-pre}%
\end{table}%
\begin{table*}[tb]
  \centering
  \resizebox{\linewidth}{!}{%
    \begin{tabular}{lrrrrrrrrr}
    \toprule
    \multicolumn{1}{c}{\multirow{2}[4]{*}{Methods}} & \multicolumn{1}{c}{MNIST} & \multicolumn{1}{c}{EMNIST} & \multicolumn{2}{c}{FMNIST} & \multicolumn{3}{c}{Cifar10} & \multicolumn{2}{c}{Cifar100} \\
\cmidrule{4-10}          & \multicolumn{1}{c}{MLP} & \multicolumn{1}{c}{MLP} & \multicolumn{1}{c}{MLP} & \multicolumn{1}{c}{Mnistnet} & \multicolumn{1}{c}{ConvNet} & \multicolumn{1}{c}{ResNet} & \multicolumn{1}{c}{RegNet} & \multicolumn{1}{c}{ResNet} & \multicolumn{1}{c}{RegNet} \\
    \midrule
    \multicolumn{9}{c}{10 Clients} \\
    \midrule
    FedAvg & 0.9017 (1.0$\times$) & 0.6108 (1.0$\times$) & 0.8183 (1.0$\times$) & 0.8573 (1.0$\times$) & 0.6153 (1.0$\times$) & 0.4759 (1.0$\times$) & 0.4498 (1.0$\times$) & 0.1575 (1.0$\times$) & 0.114 (1.0$\times$) \\
    DGC   & 0.8663 (250.0$\times$) & 0.5287 (250.0$\times$) & 0.7718 (250.0$\times$) & 0.8065 (1333.3$\times$) & 0.6151 (10.4$\times$) & 0.2113 (3571.4$\times$) & 0.3050 (757.6$\times$) & 0.0138 (3571.4$\times$) & 0.0322 (757.6$\times$) \\
    signSGD & 0.8692 (32.0$\times$) & \underline{0.5415} (32.0$\times$) & 0.7550 (32.0$\times$) & \underline{0.8198} (32.0$\times$) & 0.6180 (32.0$\times$) & \underline{0.3687} (32.0$\times$) & 0.2759 (32.0$\times$) & \underline{0.0178} (32.0$\times$) & 0.0391 (32.0$\times$) \\
    STC   & \underline{0.8848} (32.0$\times$) & 0.5258 (32.0$\times$) & \textbf{0.8016} (32.0$\times$) & \textbf{0.8427} (32.0$\times$) & \textbf{0.6187} (\textbf{32.0$\times$}) & \textbf{0.4009} (32.0$\times$) & \underline{0.3568} (32.0$\times$) & 0.0088 (32.0$\times$) & \underline{0.0416} (32.0$\times$) \\
    3SFC  & \textbf{0.8876} (\textbf{250.0$\times$}) & \textbf{0.5494} (\textbf{250.0$\times$}) & \underline{0.7881} (\textbf{250.0$\times$}) & 0.8179 (\textbf{1333.3$\times$}) & \underline{0.6182} (10.4$\times$) & 0.2567 (\textbf{3571.4$\times$}) & \textbf{0.3753} (\textbf{757.6.0$\times$}) & \textbf{0.0466} (\textbf{3571.4$\times$}) &  \textbf{0.0711} (\textbf{757.6$\times$}) \\
    \midrule
    \multicolumn{9}{c}{20 Clients} \\
    \midrule
    FedAvg & 0.9013 (1.0$\times$) & 0.6086 (1.0$\times$) & 0.8173 (1.0$\times$) & 0.8572 (1.0$\times$) & 0.6146 (1.0$\times$) & 0.4701 (1.0$\times$) & 0.4646 (1.0$\times$) & 0.1785 (1.0$\times$) & 0.1194 (1.0$\times$) \\
    DGC   & 0.8808 (250.0$\times$) & 0.5332 (250.0$\times$) & 0.7768 (250.0$\times$) & \underline{0.8207} (1333.3$\times$) & \underline{0.6115} (10.4$\times$) & 0.2542 (3571.4$\times$) & 0.3204 (757.6$\times$) & 0.0101 (3571.4$\times$) & 0.0501 (757.6$\times$) \\
    signSGD & 0.8689 (32.0$\times$) & 0.5483 (32.0$\times$) & 0.7522 (32.0$\times$) & 0.8102 (32.0$\times$) & 0.6099 (32.0$\times$) & \underline{0.3673} (32.0$\times$) & 0.3020 (32.0$\times$) & \textbf{0.0824} (32.0$\times$) & \underline{0.051} (32.0$\times$) \\
    STC   & \underline{0.8889} (32.0$\times$) & \underline{0.5512} (32.0$\times$) & \textbf{0.8020} (32.0$\times$) & 0.8198 (32.0$\times$) & \textbf{0.6125} (\textbf{32.0$\times$}) & \textbf{0.4111} (32.0$\times$) & \underline{0.3748} (32.0$\times$) & \underline{0.0734} (32.0$\times$) & 0.0499 (32.0$\times$) \\
    3SFC  & \textbf{0.8918} (\textbf{250.0$\times$}) & \textbf{0.5556} (\textbf{250.0$\times$}) & \underline{0.8013} (\textbf{250.0$\times$}) & \textbf{0.8217} (\textbf{1333.3$\times$}) & 0.6044 (10.4$\times$) & 0.3049 (\textbf{3571.4$\times$}) & \textbf{0.3854} (\textbf{757.6$\times$}) & 0.0532 (\textbf{3571.4$\times$}) & \textbf{0.0764} (\textbf{757.6$\times$}) \\
    \midrule
    \multicolumn{9}{c}{40 Clients} \\
    \midrule
    FedAvg & 0.9003 (1.0$\times$) & 0.6138 (1.0$\times$) & 0.8162 (1.0$\times$) & 0.8559 (1.0$\times$) & 0.6036 (1.0$\times$) & 0.4653 (1.0$\times$) & 0.4597 (1.0$\times$) & 0.0168 (1.0$\times$) & 0.1047 (1.0$\times$) \\
    DGC   & 0.8775 (250.0$\times$) & 0.5425 (250.0$\times$) & 0.7645 (250.0$\times$) & \underline{0.8297} (1333.3$\times$) & 0.6056 (10.4$\times$) & 0.2807 (3571.4$\times$) & 0.3379 (757.6$\times$) & 0.0094 (3571.4$\times$) & 0.0448 (757.6$\times$) \\
    signSGD & 0.8698 (32.0$\times$) & 0.5583 (32.0$\times$) & 0.7546 (32.0$\times$) & 0.8124 (32.0$\times$) & \underline{0.6102} (32.0$\times$) & \underline{0.3779} (32.0$\times$) & 0.3012 (32.0$\times$) & \underline{0.0812} (32.0$\times$) & \underline{0.0475} (32.0$\times$) \\
    STC   & \underline{0.8886} (32.0$\times$) & \textbf{0.5607} (32.0$\times$) & \textbf{0.7996} (32.0$\times$) & \textbf{0.8310} (32.0$\times$) & 0.6024 (\textbf{32.0$\times$}) & \textbf{0.4128} (32.0$\times$) & \underline{0.3603} (32.0$\times$) & \textbf{0.0818} (32.0$\times$) & 0.0414 (32.0$\times$) \\
    3SFC  & \textbf{0.8886} (\textbf{250.0$\times$}) & \underline{0.5595} (\textbf{250.0$\times$}) & \underline{0.7945} (\textbf{250.0$\times$}) & 0.827 (\textbf{1333.3$\times$}) & \textbf{0.6145} (10.4$\times$) & 0.2869 (\textbf{3571.4$\times$}) & \textbf{0.3835} (\textbf{757.6$\times$}) & 0.0560 (\textbf{3571.4$\times$}) & \textbf{0.0618} (\textbf{757.6$\times$}) \\
    \bottomrule
    \end{tabular}%
    }
    \caption{Comparison of test accuracy and compression ratio. Note that 3SFC and DGC have much higher compression ratios compared to signSGD and STC due to the limitation of quantification-based methods and the high compressing efficiency of 3SFC and DGC. Consequently, while STC seems to perform well, 3SFC achieves competing or better performance with a significantly lower communication budget. A dedicated comparison of 3SFC and STC is illustrated later to demonstrate the superiority of 3SFC in Section~\ref{sec:3sfc-stc-comp}.}
  \label{tab:accuracy-compare}%
\end{table*}%
\begin{figure}
     \centering
     \begin{subfigure}[tb]{0.23\textwidth}
         \centering
         \includegraphics[width=\textwidth]{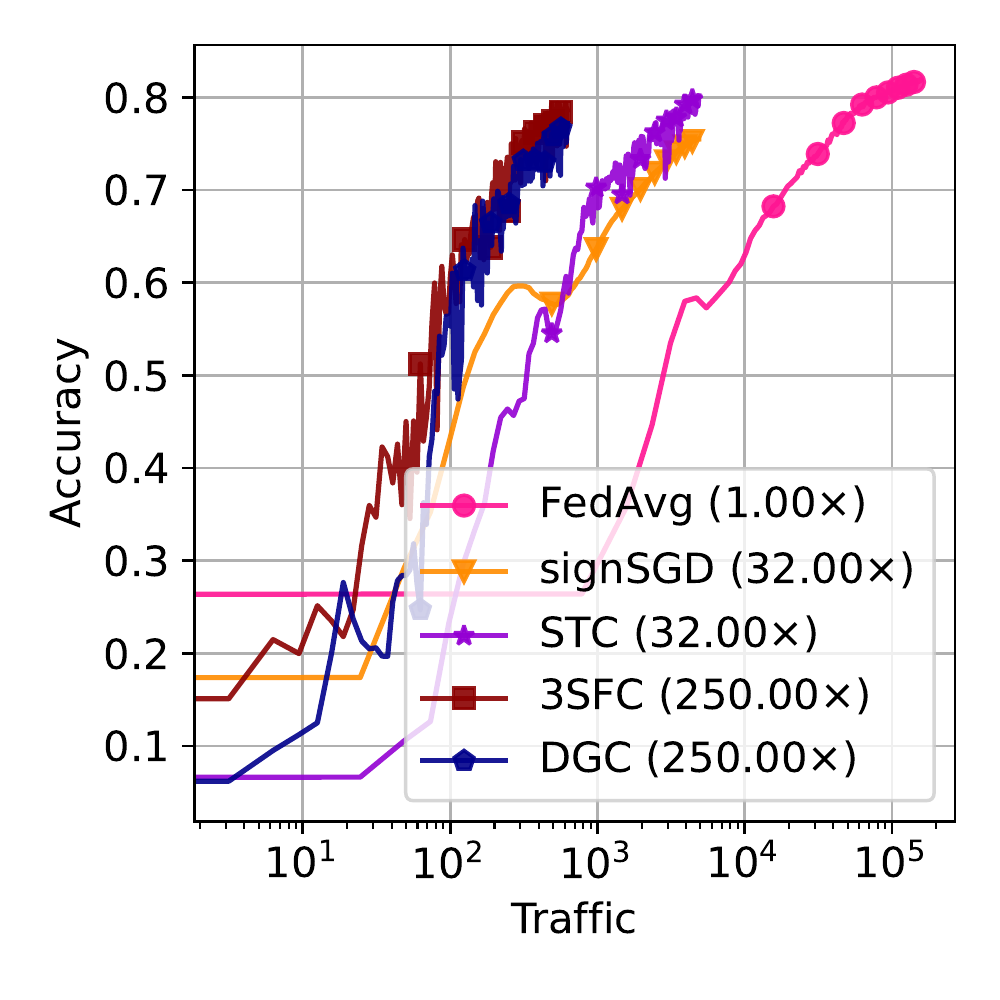}
         \caption{MLP trained on MNIST.}
     \end{subfigure}
     \hfill
     \begin{subfigure}[tb]{0.23\textwidth}
         \centering
         \includegraphics[width=\textwidth]{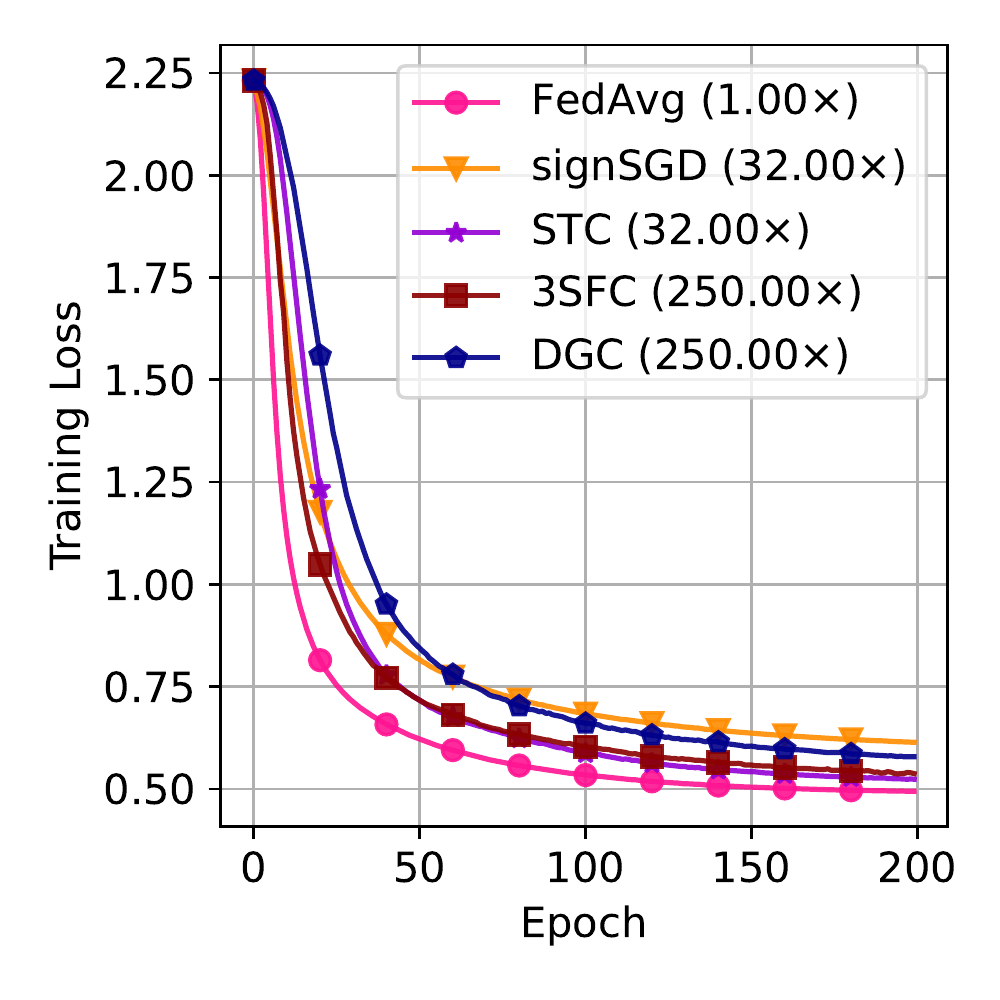}
         \caption{MLP trained on MNIST.}
     \end{subfigure}
     \\
     \begin{subfigure}[tb]{0.23\textwidth}
         \centering
         \includegraphics[width=\textwidth]{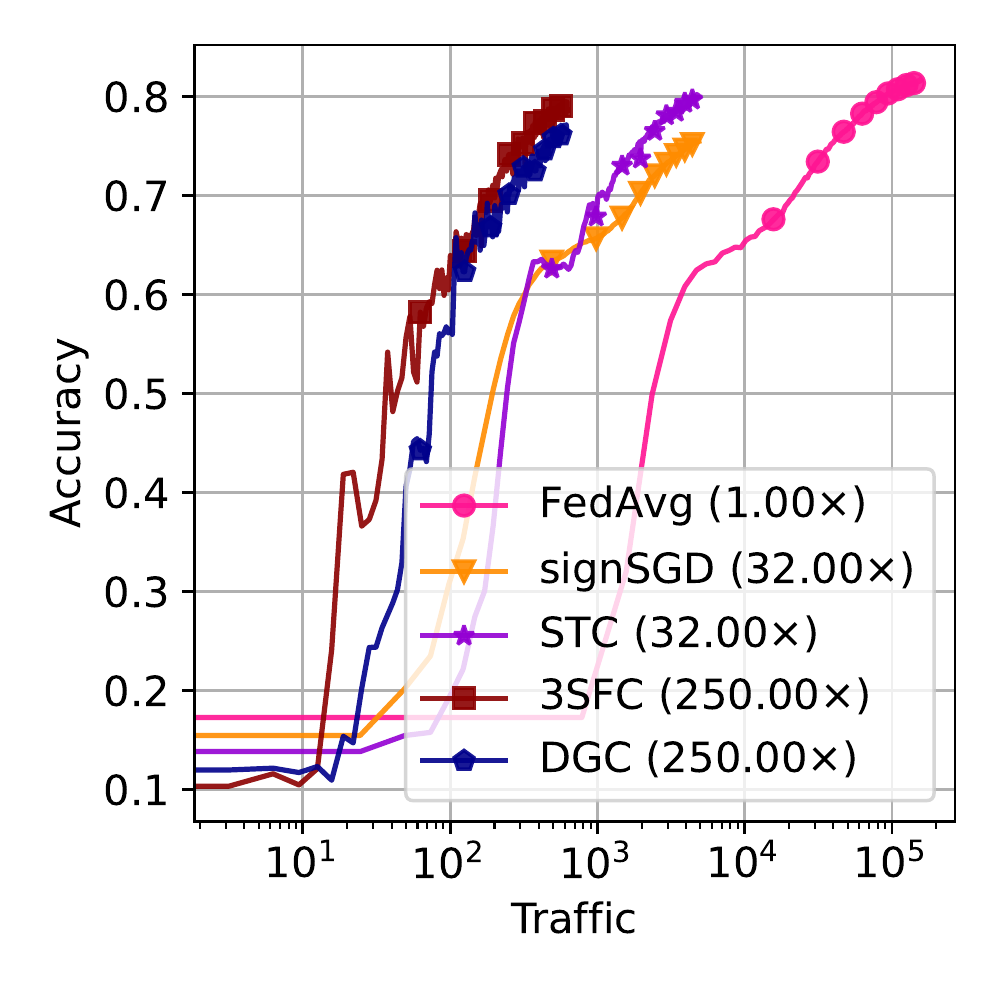}
         \caption{MLP trained on FMNIST.}
     \end{subfigure}
     \hfill
     \begin{subfigure}[tb]{0.23\textwidth}
         \centering
         \includegraphics[width=\textwidth]{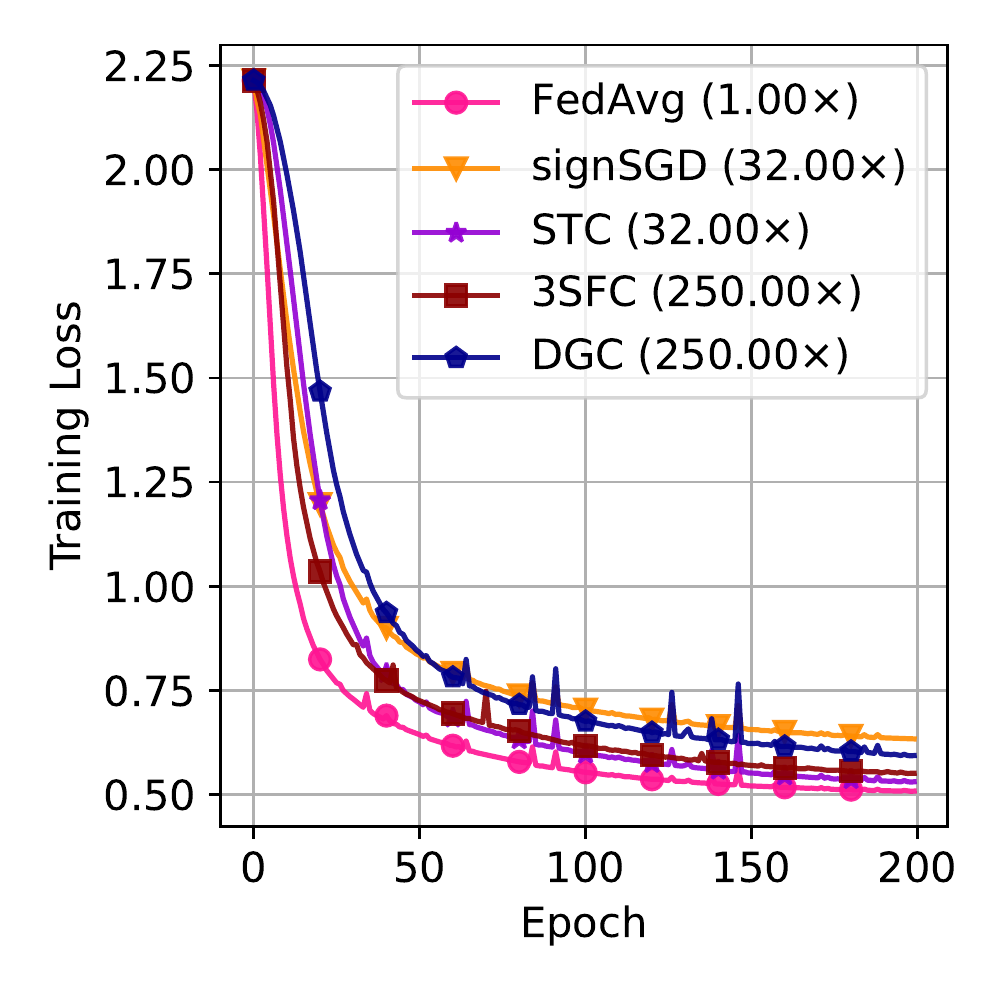}
         \caption{MLP trained on FMNIST.}
     \end{subfigure}
     \\
     \begin{subfigure}[tb]{0.23\textwidth}
         \centering
         \includegraphics[width=\textwidth]{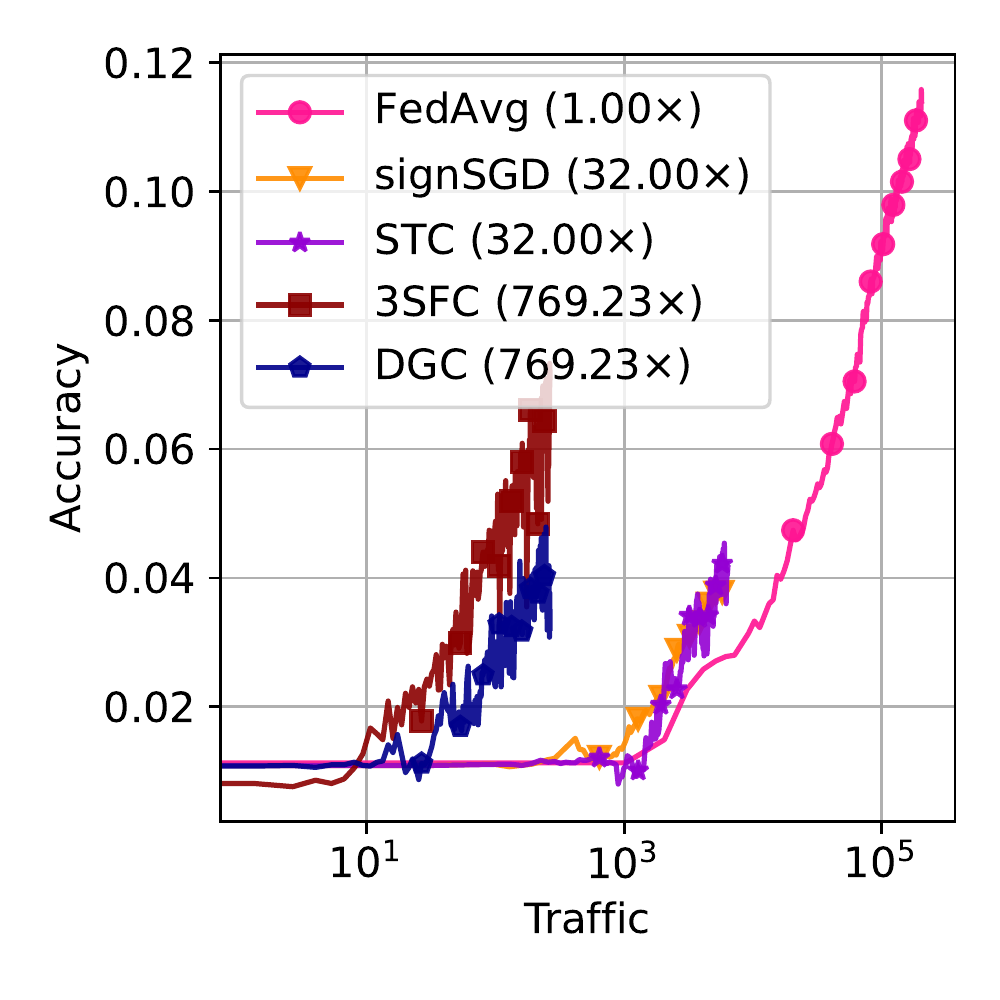}
         \caption{RegNet trained on Cifar100.}
     \end{subfigure}
     \hfill
     \begin{subfigure}[tb]{0.23\textwidth}
         \centering
         \includegraphics[width=\textwidth]{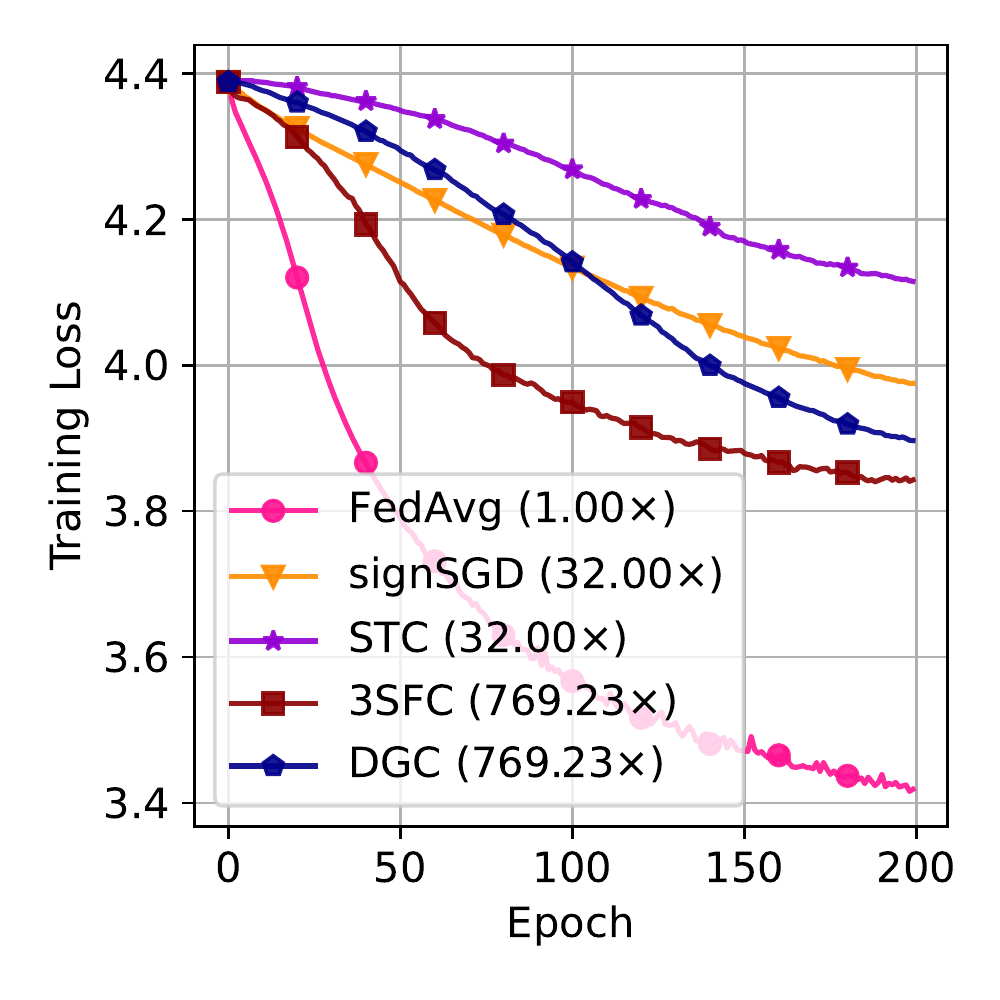}
         \caption{RegNet trained on Cifar100.}
     \end{subfigure}
        \caption{Test accuracy and training loss comparisons after 200 epochs of training. Compared to other methods, 3SFC owns the fastest convergence rate with respect to the amount of traffic communicated, with the highest compression ratio.}
        \label{fig:accuracy-loss}
\end{figure}

\textbf{Models:} To cover both simple and complicated learning problems, five models including Multi-Layer Perceptron (MLP), MnistNet, ConvNet, ResNet~\cite{he2016deep} and RegNet~\cite{radosavovic2020designing} are used in the experiments. Here, MnistNet has two convolutional layers and two linear layers, and ConvNet has four convolutional layers and one linear layer. Additionally, for ResNet and RegNet, all batch normalization layers~\cite{ioffe2015batch} and dropout layers~\cite{srivastava2014dropout} are deleted from the model as their parameters are not trainable~\cite{wang2023batch}. This simplification has also been used in previous studies\cite{sattler2019robust,zhou2021communication}.

\textbf{Competetors:} We compare 3SFC with 4 other methods: FedAvg~\cite{mcmahan2017communication}, DGC~\cite{lin2017deep}, signSGD with EF~\cite{bernstein2018signsgd} and STC~\cite{sattler2019robust}. Specifically, FedAvg is a traditional FL training method without any compression, DGC is considered as a state-of-the-art in sparsification, signSGD is a typical quantification method and STC combines sparsification and quantification (\textit{i.e.}, STC sparsifies top-$k$ parameters and quantifies the others). Note that previous work in the data distillation for FL realm (\textit{e.g.}, FedSynth~\cite{hu2022fedsynth} is considered as a state-of-the-art in data distillation for FL realm to achieve communication efficient FL) is not compared in our experiments, as it hardly converges due to the instability and collapse described in Section~\ref{sec:related-work} with high compression ratio and large datasets and models, as Table~\ref{tab:fedsynth-pre} illustrated. All later experiments are evaluated on a simulated 40 clients cluster. The CUDA version is 11.4, the Python version is 3.9.15 and the PyTorch version is 1.13.0.

\section{Analysis}
\subsection{Performance comparisons}
We first compare the final accuracy of 3SFC with other competing methods after 200 epochs of training. The learning rate is set to 0.01, the batch size is set to $256$, local iteration $K$ is set to $5$ and $\lambda$ is set to 0 for no regularization. In terms of the compression rate, we set DGC to be the same as 3SFC for all experiments for fair comparisons, because DGC is a sparsification-based method that can have an extremely low compression rate. For quantification-based methods like signSGD and STC, we leave their compression rate to be 1/32, and will later do dedicated evaluations between them and 3SFC.

The comprehensive accuracy comparison results of 3SFC and other methods are shown in Table~\ref{tab:accuracy-compare}. It can be observed that under the same compression rate, 3SFC yields higher test accuracy consistently compared to DGC after training, suggesting that 3SFC brings a faster convergence rate to the model training when the communication budget is limited. On the other hand, 3SFC still achieves comparable model performance compared to signSGD and STC, where the latter two methods communicate much more (\textit{i.e.}, 100$\times$ more for ResNet). Figure~\ref{fig:accuracy-loss} further validates the effectiveness of 3SFC by visualizing the test accuracy and training loss.
\begin{table}[tbp]
  \centering
  \resizebox{\linewidth}{!}{%
    \begin{tabular}{lrrr}
    \toprule
    Dataset+Model & \multicolumn{1}{l}{STC} & \multicolumn{1}{l}{3SFC ($2 \times B$)} & \multicolumn{1}{l}{3SFC ($4 \times B$)} \\
    \midrule
    \multicolumn{4}{c}{10 Clients} \\
    \midrule
    MNIST+MLP & 0.8848 (32.0$\times$) & \textbf{0.8961} (\textbf{125.0$\times$}) & \underline{0.8958} (\underline{62.5$\times$}) \\
    EMNIST+MLP & 0.5258 (32.0$\times$) & \underline{0.5820} (\textbf{125.0$\times$}) & \textbf{0.5955} (\underline{62.5$\times$}) \\
    FMNIST+MLP & 0.8016 (32.0$\times$) & \underline{0.8031} (\textbf{125.0$\times$}) & \textbf{0.8063} (\underline{62.5$\times$}) \\
    FMNIST+Mnistnet & \underline{0.8427} (32.0$\times$) & 0.8356 (\textbf{666.7$\times$}) & \textbf{0.843} (\underline{333.3$\times$}) \\
    Cifar10+Resnet & \textbf{0.4009} (32.0$\times$) & 0.3642 (\textbf{1785.7$\times$}) & \underline{0.3954} (\underline{892.9$\times$}) \\
    Cifar10+Regnet & 0.3568 (32.0$\times$) & \underline{0.4335} (\textbf{378.8$\times$}) & \textbf{0.4341} (\underline{189.4$\times$}) \\
    Cifar100+ResNet & 0.0088 (32.0$\times$) & \underline{0.0881} (\textbf{1785.7$\times$}) & \textbf{0.0989} (\underline{892.9$\times$}) \\
    Cifar100+RegNet & 0.0416 (32.0$\times$) & \underline{0.0946} (\textbf{384.6$\times$}) & \textbf{0.0952} (\underline{192.3$\times$}) \\
    \midrule
    \multicolumn{4}{c}{20 Clients} \\
    \midrule
    MNIST+MLP & 0.8889 (32.0$\times$) & \underline{0.8948} (\textbf{125.0$\times$}) & \textbf{0.8963} (\underline{62.5$\times$}) \\
    EMNIST+MLP & 0.5512 (32.0$\times$) & \underline{0.5832} (\textbf{125.0$\times$}) & \textbf{0.5961} (\underline{62.5$\times$}) \\
    FMNIST+MLP & 0.8020 (32.0$\times$) & \underline{0.8053} (\textbf{125.0$\times$}) & \textbf{0.8070} (\underline{62.5$\times$}) \\
    FMNIST+Mnistnet & 0.8198 (32.0$\times$) & \underline{0.8343} (\textbf{666.7$\times$}) & \textbf{0.8372} (\underline{333.3$\times$}) \\
    Cifar10+Resnet & \textbf{0.4111} (32.0$\times$) & 0.3450 (\textbf{1785.7$\times$}) & \underline{0.3654} (\underline{892.9$\times$}) \\
    Cifar10+Regnet & 0.3748 (32.0$\times$) & \underline{0.4376} (\textbf{378.8$\times$}) & \textbf{0.4508} (\underline{189.4$\times$}) \\
    Cifar100+ResNet & 0.0734 (32.0$\times$) & \underline{0.0973} (\textbf{1785.7$\times$}) & \textbf{0.1118} (\underline{892.9$\times$}) \\
    Cifar100+RegNet & 0.0499 (32.0$\times$) & \underline{0.0977} (\textbf{384.6$\times$}) & \textbf{0.1031} (\underline{192.3$\times$}) \\
    \midrule
    \multicolumn{4}{c}{40 Clients} \\
    \midrule
    MNIST+MLP & 0.8886 (32.0$\times$) & \underline{0.8932} (\textbf{125.0$\times$}) & \textbf{0.8949} (\underline{62.5$\times$}) \\
    EMNIST+MLP & 0.5607 (32.0$\times$) & \underline{0.5876} (\textbf{125.0$\times$}) & \textbf{0.5995} (\underline{62.5$\times$}) \\
    FMNIST+MLP & 0.7996 (32.0$\times$) & \underline{0.8027} (\textbf{125.0$\times$}) & \textbf{0.8073} (\underline{62.5$\times$}) \\
    FMNIST+Mnistnet & 0.8310 (32.0$\times$) & \underline{0.8374} (\textbf{666.7$\times$}) & \textbf{0.8412} (\underline{333.3$\times$}) \\
    Cifar10+Resnet & \textbf{0.4128} (32.0$\times$) & \underline{0.3747} (\textbf{1785.7$\times$}) & 0.3695 (\underline{892.9$\times$}) \\
    Cifar10+Regnet & 0.3603 (32.0$\times$) & \underline{0.4481} (\textbf{378.8$\times$}) & \textbf{0.4503} (\underline{189.4$\times$}) \\
    Cifar100+ResNet & 0.0818 (32.0$\times$) & \underline{0.1041} (\textbf{1785.7$\times$}) & \textbf{0.1189} (\underline{892.9$\times$}) \\
    Cifar100+RegNet & 0.0414 (32.0$\times$) & \underline{0.0799} (\textbf{384.6$\times$}) & \textbf{0.0889} (\underline{192.3$\times$}) \\
    \bottomrule
    \end{tabular}%
    }
  \caption{Test accuracy and compression ratio comparisons of STC and 3SFC with different communication budgets. 3SFC mostly achieves higher test accuracy while having a higher compression ratio, suggesting 3SFC compresses and decompresses the communication data more efficiently.}
  \label{tab:3sfc-stc-comp}%
\end{table}%
\begin{figure}
     \centering
     \begin{subfigure}[tb]{0.23\textwidth}
         \centering
         \includegraphics[width=\textwidth]{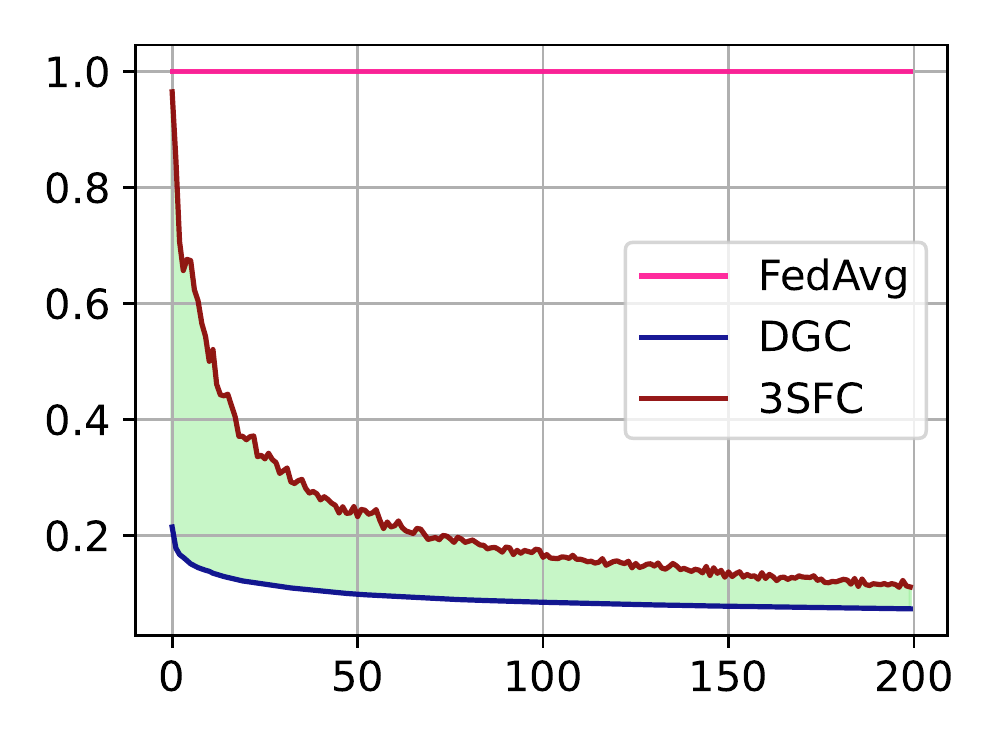}
         \caption{RegNet trained on Cifar10.}
     \end{subfigure}
     \hfill
     \begin{subfigure}[tb]{0.23\textwidth}
         \centering
         \includegraphics[width=\textwidth]{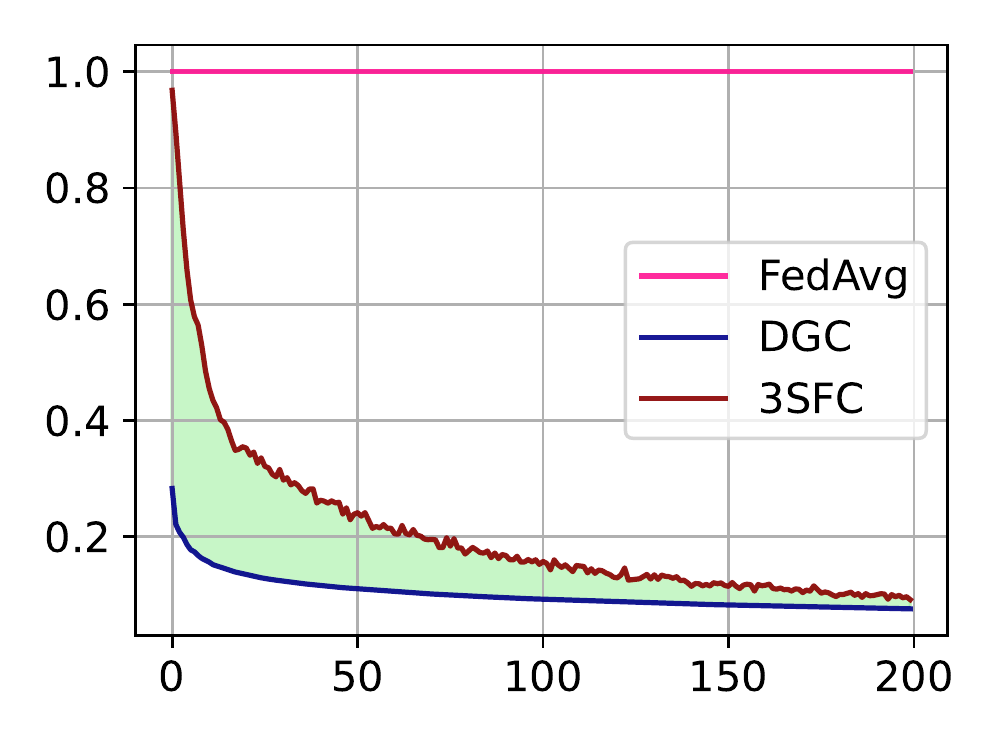}
         \caption{RegNet trained on Cifar100.}
     \end{subfigure}
     \\
     \begin{subfigure}[tb]{0.23\textwidth}
         \centering
         \includegraphics[width=\textwidth]{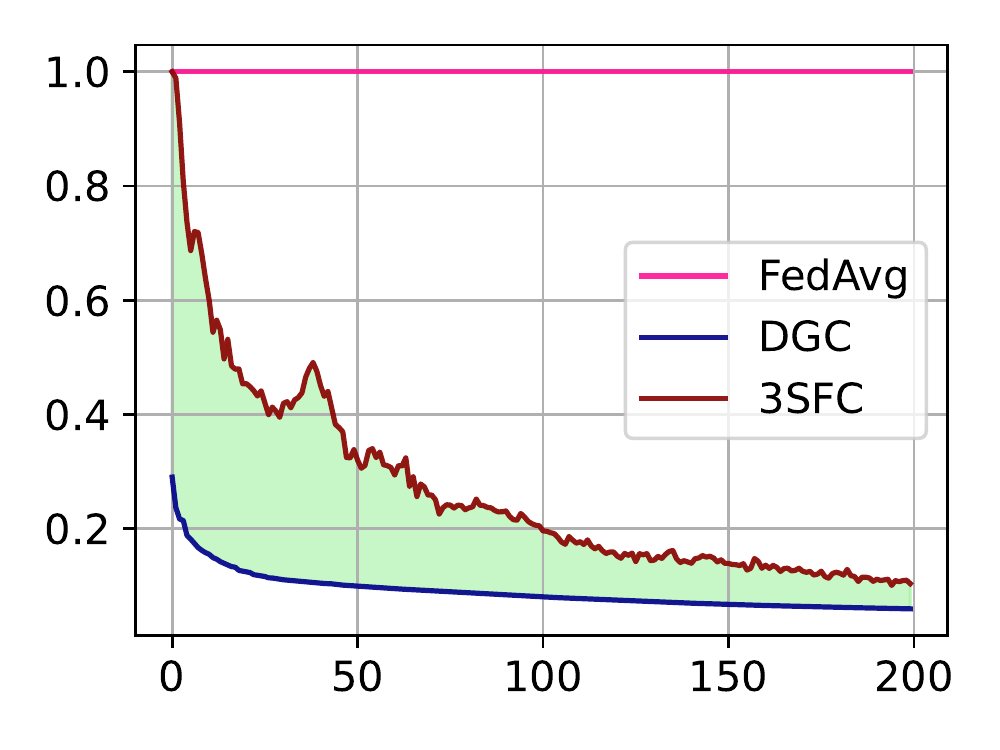}
         \caption{ResNet trained on Cifar10.}
     \end{subfigure}
     \hfill
     \begin{subfigure}[tb]{0.23\textwidth}
         \centering
         \includegraphics[width=\textwidth]{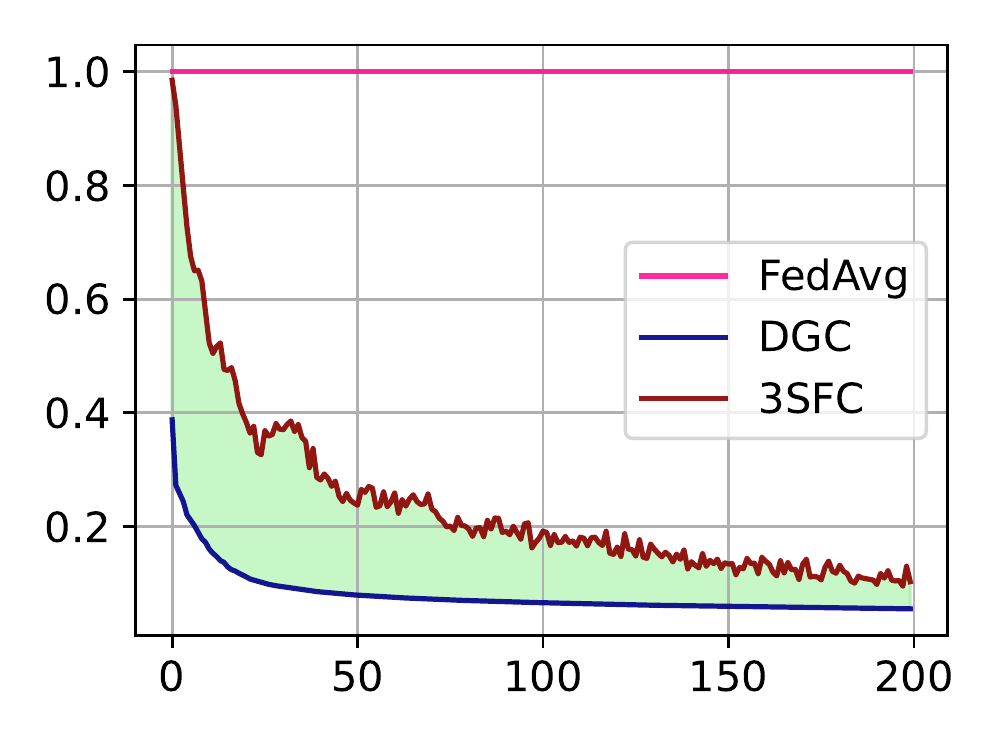}
         \caption{ResNet trained on Cifar100.}
     \end{subfigure}
        \caption{Compression efficiency comparisons. 3SFC owns significantly higher compression efficiency compared to DGC under the same compression rate, suggesting 3SFC is a much more efficient compressor compared to DGC.}
        \label{fig:compression-efficiency}
\end{figure}

\subsection{Further comparisons between 3SFC and STC}
\label{sec:3sfc-stc-comp}
To further evaluate 3SFC compared to STC for fairness, we gradually increase the communication budget of 3SFC and compare both their compression ratio and test accuracy, which is shown in Table~\ref{tab:3sfc-stc-comp}. As the table suggests, 3SFC can achieve comparable or even better test accuracy while saving a significant amount of communication traffic. For example, when training ResNet on Cifar10 with 10 clients, 3SFC reports a comparable final accuracy (0.3954 compared to 0.4009) while communicating 189.4$\times$ fewer data. On the other hand, 3SFC reaches considerably better performance for RegNet on Cifar100 (0.0946 compared to 0.0416) with 384.6$\times$ better compression ratio.

\begin{table*}[tb]
  \centering
  \resizebox{0.7\linewidth}{!}{%
    \begin{tabular}{lrrrrrrrrr}
    \toprule
    \multicolumn{1}{c}{\multirow{2}[4]{*}{Methods}} & \multicolumn{1}{c}{MNIST} & \multicolumn{1}{c}{EMNIST} & \multicolumn{2}{c}{FMNIST} & \multicolumn{3}{c}{Cifar10} & \multicolumn{2}{c}{Cifar100} \\
\cmidrule{4-10}          & \multicolumn{1}{c}{MLP} & \multicolumn{1}{c}{MLP} & \multicolumn{1}{c}{MLP} & \multicolumn{1}{c}{Mnistnet} & \multicolumn{1}{c}{ConvNet} & \multicolumn{1}{c}{ResNet} & \multicolumn{1}{c}{RegNet} & \multicolumn{1}{c}{ResNet} & \multicolumn{1}{c}{RegNet} \\
    \midrule
    \multicolumn{10}{c}{10 Clients} \\
    \midrule
    3SFC w/ EF  & 0.8876 & 0.5494 & 0.7881 & 0.8179 & 0.6182 & 0.2567 & 0.3753 & 0.3835 & 0.0711 \\
    3SFC w/o EF & 0.4580 & 0.2397 & 0.5746 & 0.7324 & 0.4495 & 0.2313 & 0.2559 & 0.0170 & 0.0235 \\
    3SFC w/ EF ($2\times B$)   & 0.8961 & 0.5820 & 0.8031 & 0.8356 & 0.6308 & 0.3642 & 0.4335 & 0.0881 & 0.0946 \\
    3SFC w/ EF ($4\times B$)   & 0.8958 & 0.5955 & 0.8063 & 0.8430 & 0.6241 & 0.3954 & 0.4341 & 0.0989 & 0.0952 \\
    3SFC w/ EF ($K=1$) & 0.6939 & 0.3152 & 0.6500  & 0.7807 & 0.5207 & 0.2001 & 0.2871 & 0.0104 & 0.0366 \\
    3SFC w/ EF ($K=10$) & 0.8961 & 0.6075 & 0.8212 & 0.8383 & 0.6333 & 0.3099 & 0.3908 & 0.0494 & 0.0778 \\
    \midrule
    \multicolumn{10}{c}{20 Clients} \\
    \midrule
    3SFC w/ EF  & 0.8918 & 0.5556 & 0.8013 & 0.8217 & 0.6044 & 0.3049 & 0.3854 & 0.0532 & 0.0764 \\
    3SFC w/o EF & 0.6707 & 0.1970 & 0.60075 & 0.7450 & 0.4625 & 0.2373 & 0.3062 & 0.0231 & 0.0295 \\
    3SFC w/ EF ($2\times B$)   & 0.8948 & 0.5832 & 0.8053 & 0.8343 & 0.6165 & 0.3450 & 0.4376 & 0.0973 & 0.0977 \\
    3SFC w/ EF ($4\times B$)   & 0.8963 & 0.5961 & 0.8070 & 0.8372 & 0.6187 & 0.3654 & 0.4508 & 0.1118 & 0.1031 \\
    3SFC w/ EF ($K=1$) & 0.7504 & 0.3387 & 0.6441 & 0.7706 & 0.5249 & 0.2155 & 0.3072 & 0.0126 & 0.0412 \\
    3SFC w/ EF ($K=10$) & 0.9063 & 0.6127 & 0.8289 & 0.8294 & 0.6049 & 0.3095 & 0.4150 & 0.0483 & 0.0831 \\
    \midrule
    \multicolumn{10}{c}{40 Clients} \\
    \midrule
    3SFC w/ EF  & 0.8886 & 0.5595 & 0.7945 & 0.8270 & 0.6145 & 0.2869 & 0.3835 & 0.0560 & 0.0618 \\
    3SFC w/o EF & 0.4830 & 0.2512 & 0.63492 & 0.7575 & 0.4742 & 0.2300  & 0.2917 & 0.0133 & 0.0277 \\
    3SFC w/ EF ($2\times B$)   & 0.8932 & 0.5876 & 0.8027 & 0.8374 & 0.6132 & 0.3747 & 0.4481 & 0.1041 & 0.0799 \\
    3SFC w/ EF ($4\times B$)   & 0.8949 & 0.5995 & 0.8073 & 0.8412 & 0.6115 & 0.3695 & 0.4503 & 0.1189 & 0.0889 \\
    3SFC w/ EF ($K=1$) & 0.6956 & 0.3129 & 0.6547 & 0.7752 & 0.5243 & 0.2254 & 0.3024 & 0.0127 & 0.0317 \\
    3SFC w/ EF ($K=10$) & 0.9072 & 0.6074 & 0.8277 & 0.8399 & 0.5875 & 0.3099 & 0.4176 & 0.0476 & 0.0748 \\
    \bottomrule
    \end{tabular}%
    }
  \caption{The ablation study with different parameters of 3SFC (\textit{i.e.}, with/without EF, communication budgets $B$, local iteration $K$). The configuration for the Base is 1 $\times B$ and $K$ = 5. From the table, it is clear that enabling EF in 3SFC has an important role in helping models converge, validating its effectiveness. Moreover, Increasing $B$ or $K$ can both further boost the convergence rate of the training.}
  \label{tab:ablation}%
\end{table*}%

\subsection{Compression efficiency}
To study why 3SFC achieves a faster convergence rate, we restrain the compression rate of 3SFC and DGC to be the same, and visualize the compression efficiency of 3SFC, DGC, and FedAvg. Here, the compression efficiency stands for how much information the compressed data carry compared to the uncompressed data. Intuitively, the compression efficiency can be represented by the $\ell_2$ distance between compressed and uncompressed data. However, since the compressed data in both 3SFC and DGC are vertical to the uncompressed data (which is illustrated by E quation~\ref{eq:s-compute}), in this subsection, we will use the cosine similarity between the compressed and uncompressed data as the compression efficiency. The visualization is shown in Figure~\ref{fig:compression-efficiency}.

In Figure~\ref{fig:compression-efficiency}, FedAvg has a constant compression efficiency of 1.0, as FedAvg does not compress the data at all. Hence, FedAvg is served as a reference here. Meanwhile, it is clear from the figure that with the same compression rate, 3SFC achieves higher compression efficiency for every communication round (\textit{i.e.}, the green area), meaning that the compression error of 3SFC is lower at every update step of the global model, contributing to the faster convergence rate of the training. Moreover, as error feedback is incorporated into both DGC and 3SFC, the compression error of each communication round will be accumulated into $\mathbf{g}_i^t$ forever. Consequently, the compression efficiency for both DGC and 3SFC decreases gradually as the training progresses.

\subsection{Ablation study}
Table~\ref{tab:ablation} shows the ablation study of 3SFC in terms of EF, communication budget $B$ and local iteration $K$. As observed from Table~\ref{tab:ablation}, compared to 3SFC w/ EF, disabling EF in 3SFC drastically degrades the model performance after training in all experiments, validating the effectiveness of EF. Moreover, MLP trained on MNIST by 3SFC w/o EF obtained a final test accuracy of 0.4580 with 10 clients (0.8876 for 3SFC w/ EF), 0.6707 with 20 clients (0.8918 for 3SFC w/ EF) and 0.4830 with 40 clients (0.8886 for 3SFC w/ EF). Such a huge performance difference in such a simple learning task effectively suggests that disabling EF also brings more instability and uncertainty to the training process. 

In terms of $B$ and $K$, when increasing $B$, the test accuracy of the model increases as well, as more data are being transferred at each communication round. On the other hand, by decreasing the local iteration $K$ from 5 to 1, the test accuracy is reduced significantly since the model has been optimized much less. Contrarily, the test accuracy boosts up when $K$ is set to 10. Consequently, increasing the communication budget $B$ is the most significant way to further boost the convergence rate of the model using 3SFC. For example, by increasing the compression rate from 0.028\% ($1\times B$) to 0.056\% ($2\times B$) for ResNet trained on Cifar100 with 40 clients, the convergence rate of the model gets doubled and the test accuracy after 200 epochs of training also increases from 0.0560 to 0.1041. However, when the communication budget is strictly limited, the convergence rate of 3SFC can be improved as well by setting a larger local iteration $K$.
\section{Conclusion}
In this paper, we propose a single-step synthetic features compressor (3SFC) for communication-efficient FL. 3SFC compresses the data using a similarity-based objective function in a single step, thus saving both compute and storage resources, and maintaining the robustness of the algorithm. Moreover, error feedback is employed to further minimize the compression error. Comparisons of test accuracy and compression ratio show that 3SFC achieves significantly faster convergence rates with lower compression rates. An ablation study demonstrates the role of different parameters, and visualizations of compression efficiency further validates the effectiveness of 3SFC.

\end{document}